\documentclass[10pt,journal,compsoc]{IEEEtran}
\ifCLASSOPTIONcompsoc
  \usepackage[nocompress]{cite}
\else
  \usepackage{cite}
\fi

\usepackage{amsmath,amssymb,amsfonts}
\usepackage{algorithmic}
\usepackage{graphicx}
\usepackage{subfigure}
\usepackage{textcomp}
\usepackage{color, colortbl}
\definecolor{whitesmoke}{rgb}{0.81,0.8,0.8}
\usepackage{multirow}
\usepackage{hyperref}
\hypersetup{
    colorlinks=true,
    linkcolor=blue,
    filecolor=magenta,      
    urlcolor=cyan,
}

\graphicspath{{imgs/}}

\begin{document}
\title{A deep convolutional neural network for classification of \textit{Aedes albopictus} mosquitoes}

\author{Gereziher~Adhane,
        Mohammad Mahdi~Dehshibi,
        and~David~Masip,~\IEEEmembership{Senior~Member,~IEEE}% <-this % stops a space
\IEEEcompsocitemizethanks{\IEEEcompsocthanksitem Department of Computer Science, Universitat Oberta de Catalunya, 08018, Barcelona, Spain.\protect\\
Corresponding author: Gereziher Adhane (e-mail: gadhane@uoc.edu).}% <-this % stops an unwanted space
\thanks{Manuscript is published.~\url{https://doi.org/10.1109/ACCESS.2021.3079700}}}

\markboth{IEEE Access}%
{Adhane \MakeLowercase{\textit{et al.}}: A deep convolutional neural network for classification of \textit{Aedes albopictus} mosquitoes}

\IEEEtitleabstractindextext{%
\begin{abstract}
Monitoring the spread of disease-carrying mosquitoes is a first and necessary step to control severe diseases such as dengue, chikungunya, Zika or yellow fever. Previous citizen science projects have been able to obtain large image datasets with linked geo-tracking information. As the number of international collaborators grows, the manual annotation by expert entomologists of the large amount of data gathered by these users becomes too time demanding and unscalable, posing a strong need for automated classification of mosquito species from images. We introduce the application of two Deep Convolutional Neural Networks in a comparative study to automate this classification task. We use the transfer learning principle to train two state-of-the-art architectures on the data provided by the Mosquito Alert project, obtaining testing accuracy of 94\%. In addition, we applied explainable models based on the Grad-CAM algorithm to visualise the most discriminant regions of the classified images, which coincide with the white band stripes located at the legs, abdomen, and thorax of  mosquitoes of the \textit{Aedes albopictus} species. The model allows us to further analyse the classification errors. Visual Grad-CAM models show that they are linked to poor acquisition conditions and strong image occlusions.
\end{abstract}

}

\maketitle
\IEEEraisesectionheading{\section{Introduction}\label{sec:introduction}}
\IEEEPARstart{D}{eveloping} automated tools for identifying and classifying mosquito species can be beneficial to entomologists who study the life cycle, distribution, ecology, behaviour, and population dynamics of mosquitoes in order to minimise or monitor the spread of mosquito-borne diseases~\cite{01}. Conventional approaches used in ecology are highly dependent on human expertise, spectral analysis of wingbeat waveforms~\cite{16ouyang2015mosquito,17li2005automated} and larval DNA analysis~\cite{18walton1999molecular}. Since these approaches cannot be commonly used to study larger populations of mosquitoes, the development of automated tools to classify mosquito species is demanded. The vision-based classification of mosquito species has recently been used in several studies~\cite{19okayasu2019vision,20lorenz2015artificial,21kesavaraju2012new,22sanchez2017mosquito,36park2020classification} and shown promising results when applied to images captured in laboratory settings. However, the development of tools trained with a wide range of images captured in uncontrolled conditions (i.e., taken by volunteers using smartphones) could enable the ecologist to identify larger numbers of species of mosquitoes to address environmental concerns caused by the spread of these species~\cite{52pataki2021deep}.

The Mosquito Alert citizen science platform~\cite{01} was launched in 2014 to help combating the spread of disease-carrying mosquitoes in Spain. The purpose of this project, coordinated by a team in a small network of academic institutions in Spain (CEAB-CSIC, UPF, CREAF, and ICREA), is to rise awareness among the population and build expert-validated citizen science networks to monitor and control disease-carrying mosquitoes. The platform brings together citizens, entomologists, public health authorities and mosquito control services to help reduce mosquito-borne diseases in Spain, and is currently expanding and adjusting its tools for combating disease-carrying mosquitoes across Europe and worldwide. 

The Asian tiger mosquito (\textit{Aedes albopictus}) is one of the first disease-carrying species that was targeted in the platform. This species is invasive in Europe and landed in Spain in 2004 near Barcelona, where it currently resides. Since then, it has invaded all of the eastern coasts of Spain and has rapidly invaded the the southern and western (inland) regions of the Iberian Peninsula. 

The data collection method for this platform is mainly based on images of tiger mosquitoes and mosquito breeding sites submitted by residents. The submitted images are then inspected, each validated and classified through visual inspection by a team of three entomologists, along with another expert in the event of a dispute, to be included in the database and the public online map with a classification label. However, the potential increase in the number of images submitted as the project expands makes the hand-crafted inspection of the images one of the scalability challenges of the project. The team has therefore already begun to develop automated tools to obtain integrated and real-time information on the spread of mosquitoes in order to promote the development of preventive strategies to combat various mosquito-borne diseases~\cite{52pataki2021deep}.

\begin{figure*}[!htbp]
    \xdef\xfigwd{\columnwidth}
    \centering
    \includegraphics[width=0.8\linewidth]{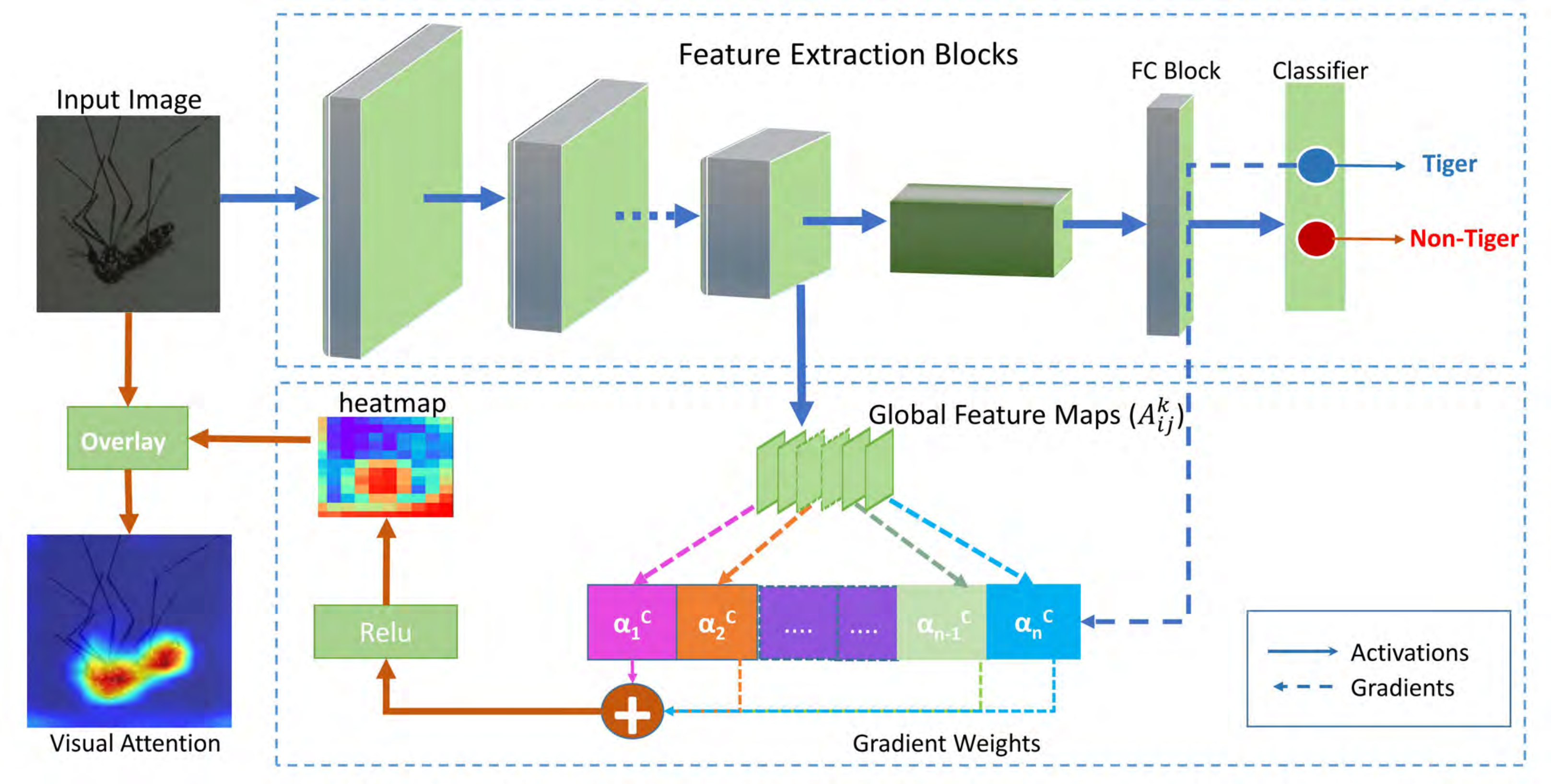}
    \caption{The pipeline of the proposed methodology for the classification of in-the-wild tiger mosquitoes species, where the Grad-CAM component component gives fine-grained explanations of the predicted classes.}
    \label{fig:architecture}
\end{figure*}

Following~\cite{52pataki2021deep}, we proposed an explainable deep learning architecture to classify \textit{Aedes albopictus} from other mosquito species. More specifically, inspired by the VGG architecture~\cite{13simonyan2014very} and the Grad-CAM visualisation~\cite{02selvaraju2016grad}, we proposed to use a deep convolutional neural network (CNN) in which the CNN component extracts discriminative features, and the Grad-CAM explains the CNN learnt features (see Figure~\ref{fig:architecture}). We used the concept of transfer learning to address small sample sizes in the target domain. Our architecture was then adapted from the ImageNet domain by fine-tuning with 6378 (tiger and non-tiger mosquitoes) samples in the target domain. Deep learning models are usually trained using millions of samples. When we compare this scale to the total number of images in the Mosquito Alert project, we can consider this classification task as a sort of small sample size problem. The Grad-CAM visualisation component gives fine-grained explanations of the predicted classes to explain to researchers or entomologists how the proposed architecture sees or perceives \textit{Aedes albopictus} species. The accuracy obtained (close to $94\%$) are consistent with \cite{52pataki2021deep} (where ResNet 50 was used as a decision rule instead of VGG). 

The explainability allows us to examine the regions that the model uses to distinguish the tiger mosquitoes. It also enables us to extract the relevant cues for evaluating classification errors and fine-tuning the architecture's parameters. In addition, the visualisation component allows entomologists to identify the key mosquitoes' body regions used by deep learning to classify the associated classes. In our experiments, we compared VGG16~\cite{13simonyan2014very} and ResNet50~\cite{12he2016deep} architectures to show the effect of fine-tuning on samples belonging to the target domain. We also determined the morphological characteristics of the \textit{Aedes albopictus} species selected by CNNs to show the superiority of the proposed methodology compared to entomologists, where the input images are not of sufficient quality for visual inspection.

The rest of this paper is organised as follows: Section~\ref{sec:survey} surveys the previous studies. The proposed methodology is detailed in Section~\ref{sec:methodology}. Section~\ref{sec:experiments} presents experiments results. Finally, Section~\ref{sec:conclusion} concludes the paper.

\section{Related Work}\label{sec:survey}
Rapid and accurate identification of insects has important applications like detection of disease vectors and understanding of biological diversity. Due to the variety of species and similarities between families of the same species, insect identification has always been a challenging task for entomologists and taxonomists. Many studies have proposed automated systems~\cite{40mukundarajan2017using,41wilke2016morphometric,16ouyang2015mosquito,17li2005automated, 42feng2016software,43martineau2017survey,44yang2015tool,37de2016detection,38fuchida2017vision,39chen2014flying} to identify and classify insects. Among different species of insects, the identification of mosquitoes is important due to their impact on the outbreaks of infectious diseases such as yellow fever, dengue, Zika and chikungunya. Strategies for automatic mosquito detection can be categorised into audio and visual feature analysis categories. In order to analyse the audio features, researchers mainly used acoustic wingbeat recordings~\cite{40mukundarajan2017using,41wilke2016morphometric,16ouyang2015mosquito,17li2005automated}. For instance, Cator et al.~\cite{47cator2009harmonic,48cator2011behavioral} used optical sensors to record the wingbeat audio of the \textit{Aedes aegypti} mosquito species. They observed that the male and female species shifted their wingbeat frequency to match during mating.

Recently, convolutional neural networks (CNN) achieved expert-level performance in different tasks, ranging from image analysis~\cite{krizhevsky2017imagenet} to health risk assessment~\cite{zeleznik2021deep}. Several studies have trained different CNN architectures on small image datasets captured in laboratory settings to track, identify and classify mosquitoes~\cite{19okayasu2019vision,22sanchez2017mosquito}. 

Schreiber et al.~\cite{51schreiber2020detecting} used CNN to classify adult \textit{Aedes aegypti} mosquito species by analysing wingbeat audio recorded by smartphones. They trained a binary, multiclass and ensemble of binary classifiers with the recording spectrogram to represent mosquito wingbeat frequency over time. Since wingbeat recordings were performed in an environment with minimal background noise, the efficiency of classifiers in the processing of recordings in a noisy environment may be reduced. In addition, this method is expensive due to the expense of optical sensors and the difficulty of obtaining high-quality audio recordings.

Faud et al.~\cite{49fuad2018training} and Ortiz et al.~\cite{22sanchez2017mosquito} suggested CNN-based mosquito classification techniques for larvae images. They fine-tuned pre-trained CNNs with a small dataset and achieved high classification accuracy. Their approaches can identify whether a mosquito from larvae is a vector. Since data acquisition and experiments were carried out in the laboratory setting, the use of these tools by taxonomists and health workers in the field is not feasible. Pre-trained CNN architectures were investigated in~\cite{36park2020classification} to identify six different species of mosquitoes with similar morphological structure. On a small sample of data set, they used transfer learning to train the proposed architecture. They achieved high classification accuracy and could locate each mosquito species' discriminative regions. However, the images were gathered in a laboratory setting, and data augmentation technique was used to increase the number of samples. As a result, the discriminative regions for some of the mosquito images could needlessly be transformed, resulting in inaccurate identification of the species of mosquitoes.

Motta et al.~\cite{50motta2019application} implemented an automated morphological classification tool using LeNet, AlexNet and GoogLeNet to facilitate the automatic classification of adult mosquitoes. The main objective was to develop a tool that entomologists and health workers can use in real-world settings to encourage volunteers to participate in controlling the vector-borne disease. Although they reported fine-grained classification accuracy, the generalisability of these methods for the classification of data acquired in-the-wild is not remarkable.

Rodriguez et al.~\cite{28rodriguez2016machine} have created an application to facilitate the verification of submitted photos by citizens for the Alert project. This application could verify a portion of the collected reports which enabled entomologists to focus on larger reports. They compared the performance of Naive Bayes, K-nearest neighbour, decision tree and random forest classifiers when applied to features such as date of submission, location and user history (previous valid reports, submission frequency, accuracy, and mobility).

Pataki et al.~\cite{52pataki2021deep} trained a deep learning model using the mosquito alert~\cite{01} data set and looked at several aspects of the data collected from citizens, such as the quality, geographical diversity and number and usefulness of the submitted images. The quality and usefulness of citizen-submitted mosquito images were evaluated over time and place. They also looked at the current regional distribution and dynamics of invasive species, which is a practical way to keep track of the spread of vectors in colonised territories.

\begin{figure}[!htbp]
    \xdef\xfigwd{\columnwidth}
    \centering
    \includegraphics[width=0.6\linewidth]{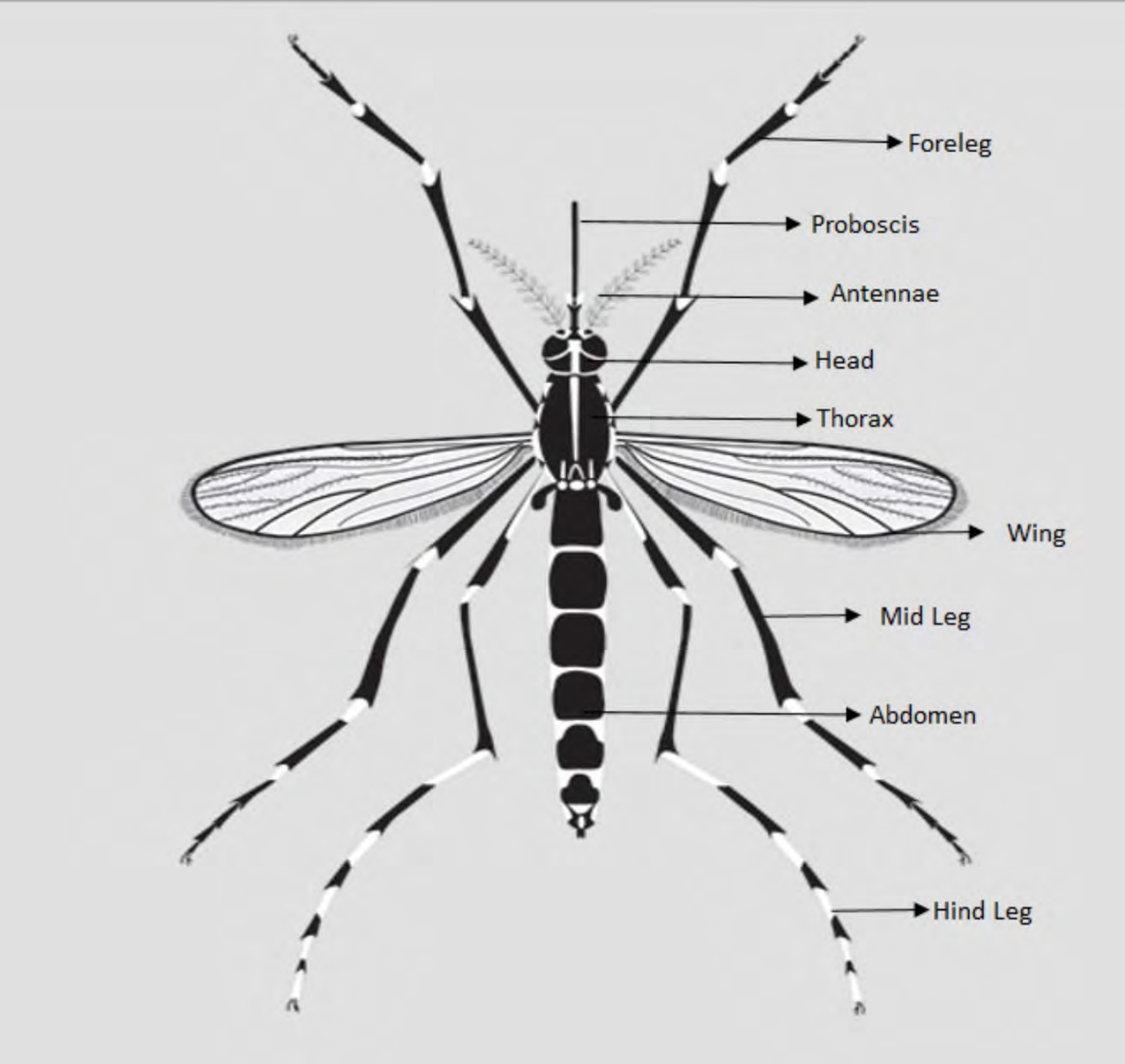}
    \caption{Anatomy of \textit{Aedes albopictus} mosquito. This image is taken from \url{https://us.biogents.com/aedes-albopictus-asian-tiger-mosquito/}.}  
    \label{fig:anatomy}
\end{figure}

Motivated by~\cite{36park2020classification,50motta2019application,52pataki2021deep} and owning to the availability and pervasive usage of smartphones that allow volunteers to send pictures of mosquitoes species, we fine-tuned the pre-trained VGG16~\cite{13simonyan2014very} on ImageNet~\cite{russakovsky2015imagenet} to classify \textit{Aedes albopictus} mosquito species. To fine-tune the model, we used images taken with smartphones and digital cameras by citizen scientists participating in the Mosquito Alert platform. The experimental results show 94\% accuracy in the classification of tiger mosquito species, where the morphological characteristics inspected by the networks are consistent with the visual elements used by expert annotators. The use of Grad-CAM~\cite{02selvaraju2016grad} has shown that the key components of tiger mosquito specimens that support network convergence are the white band on the legs, abdomen patches, head, and thorax, see Figure~\ref{fig:anatomy} for the anatomy of \textit{Aedes albopictus}. Details of the proposed methodology are given in the following section.

\section{Methodology}\label{sec:methodology}
We formulate this task as a binary classification problem in which the response $\mathbf{Y} \in \{0,1\}$ is predicted given the input image $\mathbf{X} \in \mathbb{R}^{C \times H \times W}$ with number of channels $C$, height $H$ and width $W$, and the loss is measured by the cross-entropy. Given a dataset of $n$ mosquito images $\mathcal{D} = \{(x_{i}, y_{i})\}_{i=1}^{n}$, our goal is to learn a neural network $\mathcal{F}(x) = \mathbb{E}[\mathbf{Y}|\mathbf{X} = x]$ that minimises the loss function in a way that $y_{i} = 1$ if the $i$-th sample is \textit{Aedes albopictus}, and $y_{i} =0 $ otherwise. Instead of using a random weight initialisation, we used a pre-trained VGG16 ($\mathcal{F}$)~\cite{13simonyan2014very} with ImageNet~\cite{russakovsky2015imagenet} as a starting point for learning rich representations of features. We used the transfer learning concept to fine-tune $\mathcal{F}$ to the mosquito classification task, providing a faster and easier convergence while using a limited number of training images. Since the input size of $\mathcal{F}$ is $3 \times 224 \times 224$, we need to scale $H$ and $W$ in our dataset. In addition to scaling, we normalise input images using $z$-score method (see Eq.~\ref{eq:01}) with which the dataset has mean and standard deviation values of 0 and 1, respectively.

\begin{equation}
    \label{eq:01}
    \begin{matrix}
        z = \frac{x-\bar{X}}{\sigma}, \\ 
        \bar{X} = \frac{1}{n}\sum_{i=1}^{n}x_{i},~~\sigma = \sqrt{\frac{1}{n}\sum_{i=1}^{n}(x_{i}-\bar{X})^{2}}.
\end{matrix}
\end{equation}
where $\bar{X}$ and $\sigma$ are the mean and standard deviation of dataset. The modified VGG16 architecture consists of 13 convolution layers with stride $1 \times 1$ and padding $[1,  1,  1,  1]$. Each convolution layer follows by a max-pooling and rectifier linear unit (ReLU). The architecture tail contains two fully connected, ReLU and dropout layers, which are serially connected. Finally, there is a softmax module which is followed by a cross-entropy loss function to map the feature vector into \textit{tiger} or \textit{non-tiger mosquito} classes. The pipeline of the architecture is presented in Figure~\ref{fig:architecture}.

The softmax also calculates the gradients of the identified class that is used in the backpropagation phase. This gradient generates the heatmaps of the convolutional feature maps. The neuron importance weights ($\alpha$) are given by the  winning class gradients and feature maps of the last convolutional layer. We used Grad-CAM~\cite{02selvaraju2016grad} visualisation algorithm to compare the gradient of a predicted class with its final convolutional layer and to weight it against the corresponding class. This module helps to generate explanation by using the preserved spatial information in the convolutional layers during training phase and generate a heatmap visualisation that focus on the region of interest (ROI) with a higher resolution.

Gradient-based Class Activation Maps (Grad-CAMs) are generated from the gradient score of the predicted class $\tilde{y} = \mathcal{F}(x')$ with respect to the feature map of the last convolution layer. More formally, for resized input image $x'$, let $A_{k}(i,j)$ denote the $k$-th activation unit in the last convolutional layer, the neuron importance weight corresponding to class $c$ for unit $k$ is given by Eq.~\ref{eq:02}

\begin{equation}
    \label{eq:02}
    \alpha_{k}^{c}(i,j) = \frac{1}{{H'}\times{W'}}\sum_{i=1}^{H'}\sum_{j=1}^{W'}\frac{\partial \tilde{y}^{c}}{\partial A_{k}(i,j)}.
\end{equation}
where $\alpha$ describes the importance of visual patterns at different spatial locations for a given class $c$. Double summation in Eq.~\ref{eq:02} resemblances the global average pooling over height and width to obtain a feature map of shape $[k, 1, 1]$. To generate the heatmap $L_{Grad-CAM}$, in which the ROI is highlighted, we apply ReLU activation function to the weighted sum of the alpha values as in Eq.~\ref{eq:03}.

\begin{equation}
    \label{eq:03}
    L_{Grad-CAM}^{c} = ReLU\left ( \sum_{k}\alpha_{k}^{c} A_{k} \right )
\end{equation}

\begin{figure}[!htbp]
    \centering
    \includegraphics[width=0.8\linewidth]{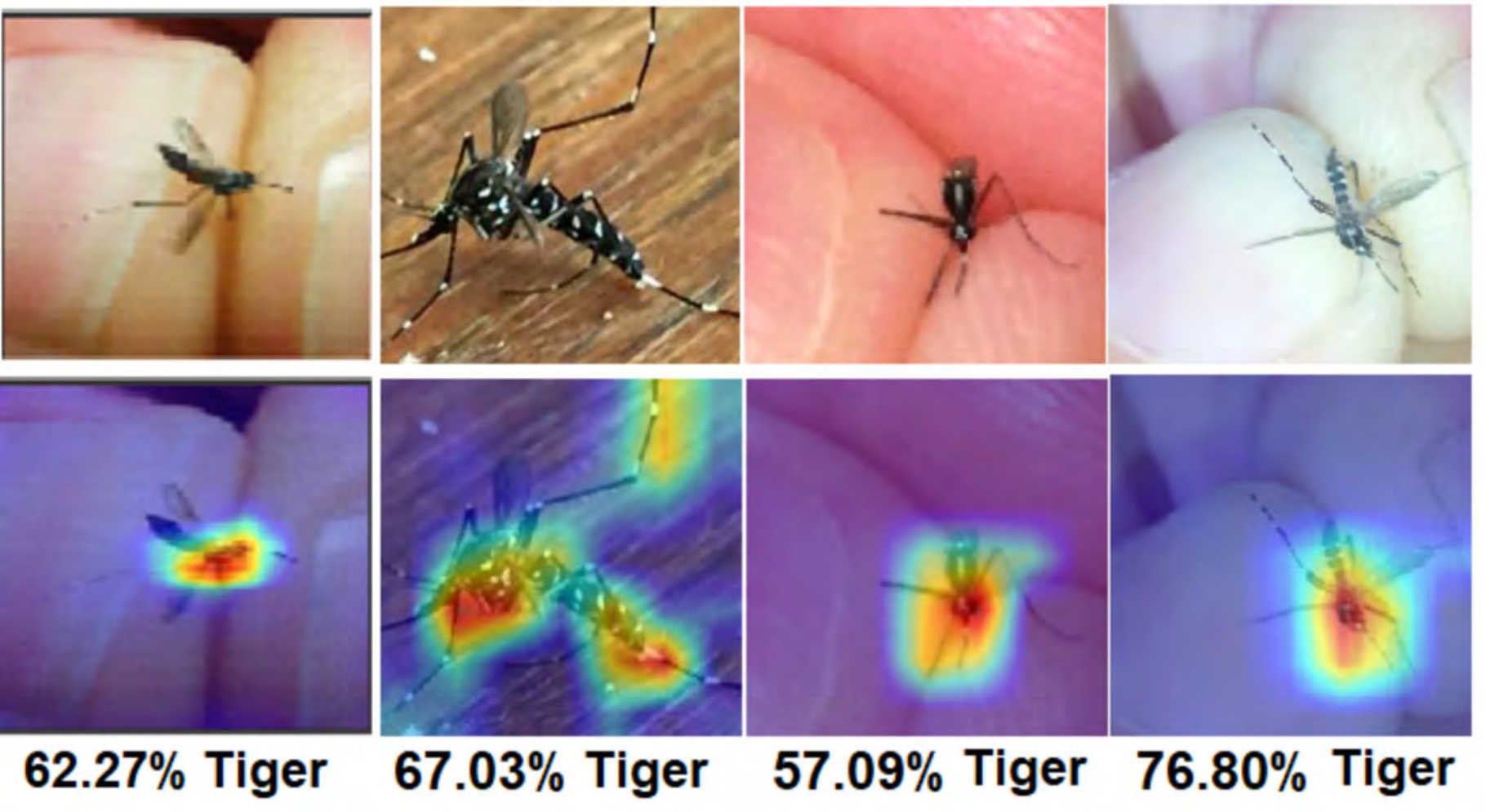}
    \caption{Examples of tiger mosquito images' heatmap visualisation from the last convolutional layer.}  
    \label{fig:heatmap}
\end{figure}

As a result of applying global average pooling and ReLU activation function, the size of the heatmap is smaller than the original input image. The generated heatmap must therefore be up-sampled to fit the size of the original input image. Figure~\ref{fig:heatmap} shows four samples of the dataset along with the generated heatmap.

\section{Experiments} \label{sec:experiments}
We evaluate our methodology in three experiments: (1) training and validation of the deep architecture on data set annotated by expert entomologists; (2) training the model on data annotated by expert entomologists and testing it with data submitted by volunteers, i.e., tiger mosquitoes with not-classified confidence level; and (3) explainability. We present the experiments' characteristics by providing descriptions of the data set, the architecture and the visual analysis of ROIs using Grad-CAM.

\subsection{Data set}
The Mosquito Alert platform~\cite{01} to monitor and control diseases-carrying  mosquitoes is open sourced\footnote{The curated database with the images can be found at~\url{http://www.mosquitoalert.com/en/mosquito-images-dataset}}. The data collection process for this platform relies on images mosquitoes and mosquito breeding sites submitted by volunteers. The photos submitted are then inspected, validated and classified by a team of three expert entomologists and, in case of doubt, by a super-expert assigning a final classification, in coordination with the other three experts. 
%%%PLACEMENT
\begin{figure*}[!htbp]
    \centering
    \includegraphics[width=0.7\linewidth]{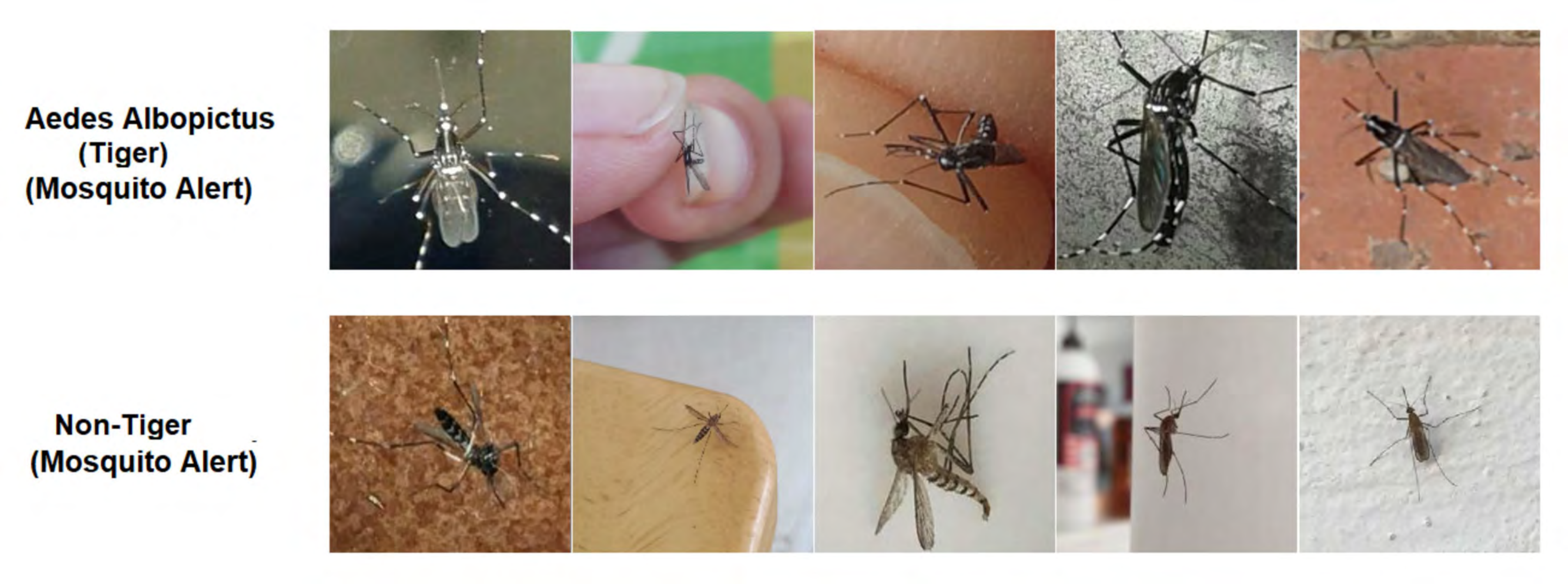}
    \caption{Sample of \textit{tiger} [first row] and \textit{non-tiger} [second row] mosquitoes from the Mosquito Alert data set.}
    \label{fig:samples}
\end{figure*}

Each image is categorised into one of four classes (\textit{Aedes albopictus}, \textit{Aedes aegypti}, other species, can not tell). The experts mark each image's classification as confirmed, probable, or not-classified in each category. In this work, we used 3364 images labelled as confirmed \textit{Aedes albopictus} cases as positive samples (i.e., tiger), and 3014 \textit{Aedes aegypti} and other species as negative samples, i.e., non-tiger. As a result, the architecture is trained using a total number of 6378 tiger and non-tiger images. Figure~\ref{fig:samples} shows five tiger and five other mosquito species samples collected in the wild.

\subsection{Architecture details and evaluation metrics}
For a fair comparison, the following hyperparameters were used in all classification models. Both VGG16~\cite{13simonyan2014very} and ResNet50~\cite{12he2016deep} architectures were pre-trained on ImageNet~\cite{russakovsky2015imagenet} with input size of $H', W' = 224$. All models were trained end-to-end in PyTorch~\cite{14NEURIPS2019_9015} using the Tesla K80 GPU, where the implementations are available at\footnote{The implementation is available at:~\url{https://github.com/ageryw/MosquitoClassification}}. In the training of the two architectures, stochastic gradient descent with momentum algorithm~~\cite{14NEURIPS2019_9015} was used to update learning parameters with initial learning rate and momentum of 0.001 and 0.7, respectively. The learning rate decays by a factor of $gamma = 0.1$ every 7 epochs using StepLR scheduler. We fed networks with a mini-batch size of 64 and the optimisation stopped after 25 epochs. In the experiments, we did not use data augmentation to avoid unrealistic changes in micro-morphological patterns of mosquitoes' body that could have skewed the final results. We also checked the validation in each epoch to manage the small sample size.

We used the accuracy and loss of classification to measure the performance of the model. To provide more insight into the performance of predictive model, we calculated precision, recall, and F1, see Eq.~\ref{eq:04}.

\begin{align}
    \label{eq:04}
    Precision = \frac{TP}{TP + FP}, ~~~~ Recall = \frac{TP}{TP + FN}, \nonumber \\
    F1 = 2\times \frac{Precision \times Recall}{Precision + Recall}.
\end{align}
where $TP$, $FP$ and $FN$ stand for true positive, false positive and false negative, respectively.

\subsection{Experimental results}
In our experiments, we first assess the efficiency of the proposed approach in annotated cases by expert entomologists. This is then compared to the output of the model in classifying images that have been tagged as not-classified. In addition, we randomly selected 150 mosquito images (clear and damaged images of \textit{tiger} and \textit{non-tiger}) to observe the predictive scores concerning their visualisation in order to explain the main sources of CNN-induced errors.

\begin{figure}[!htbp]
    \centering
    \subfigure[]{\includegraphics[width=1\linewidth]{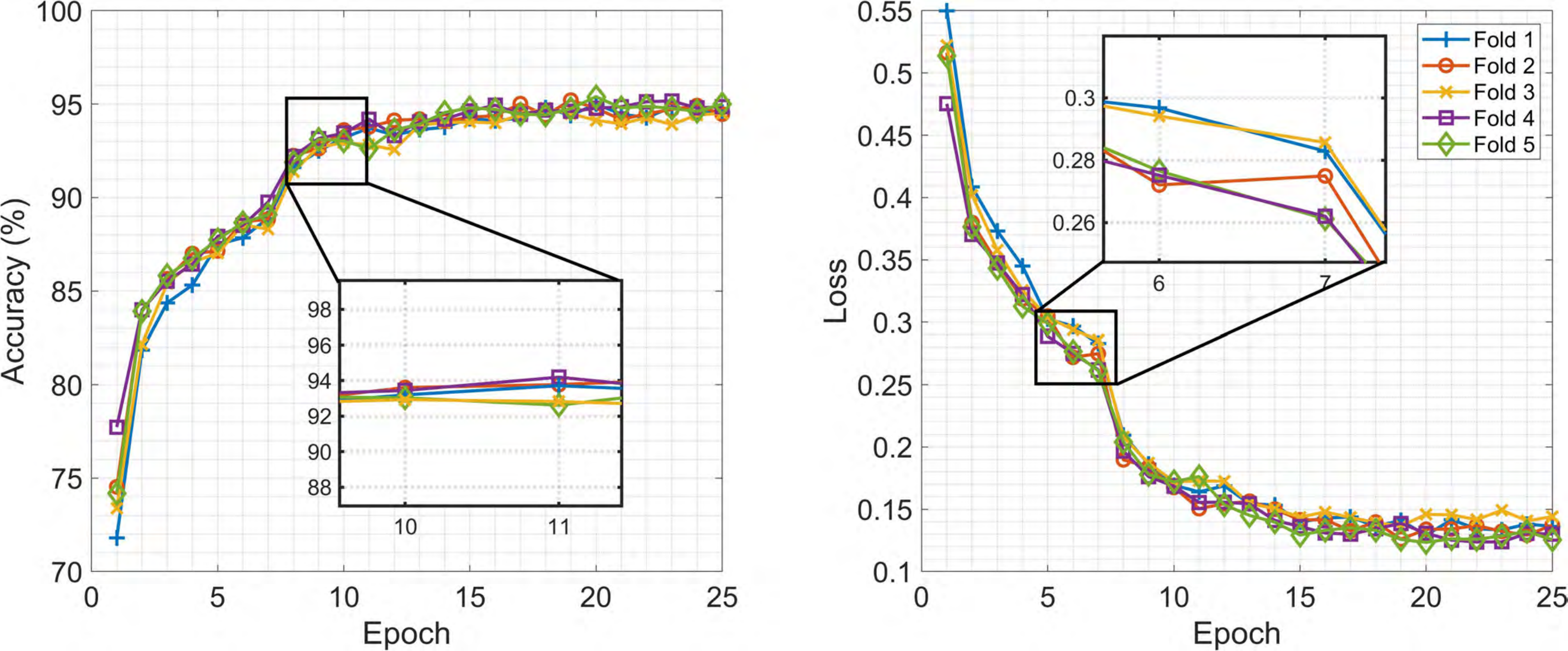}}
    \subfigure[]{\includegraphics[width=1\linewidth]{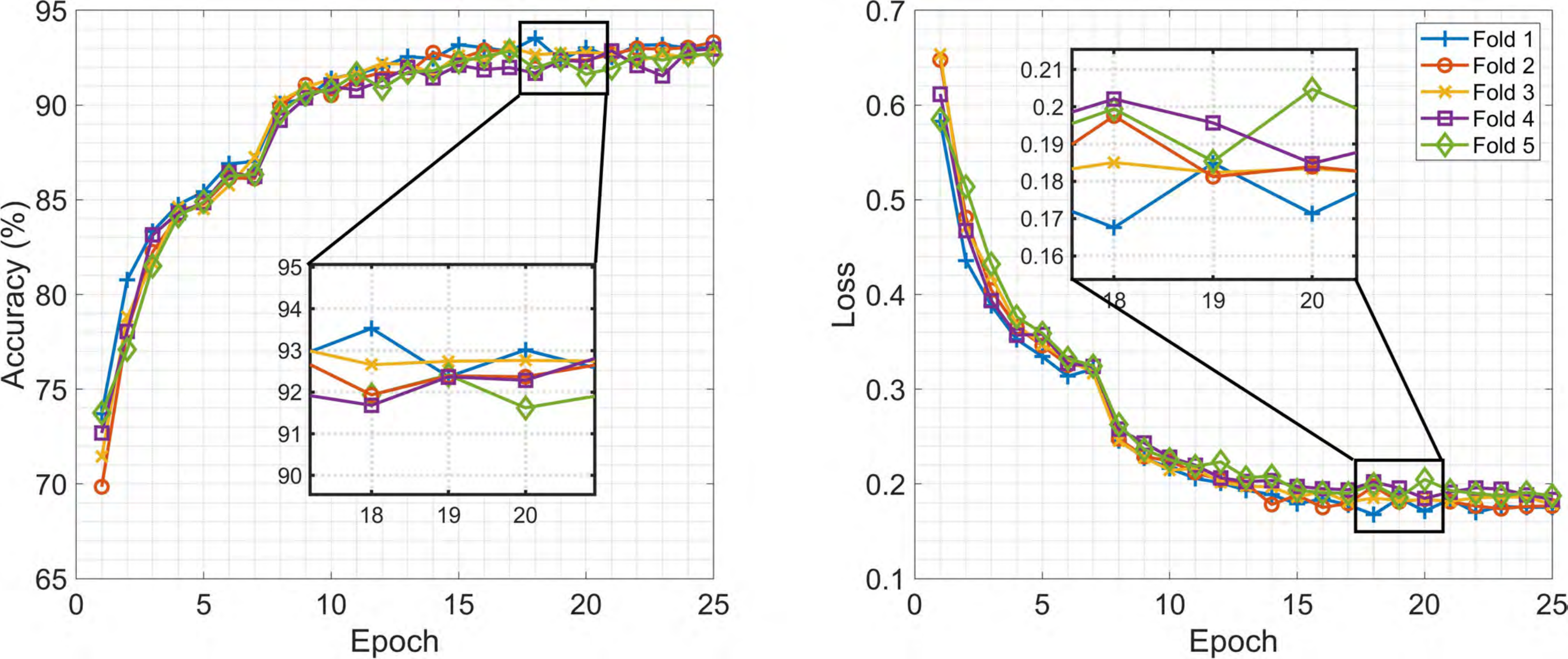}}
    \caption{Training progress in the accuracy and loss of (a) VGG16 and (b) ResNet50. Each plot contains progress for 5 folds.}
    \label{fig:trainingAccLoss}
\end{figure}

\begin{figure}[!htbp]
    \centering
    \includegraphics[width=1\linewidth]{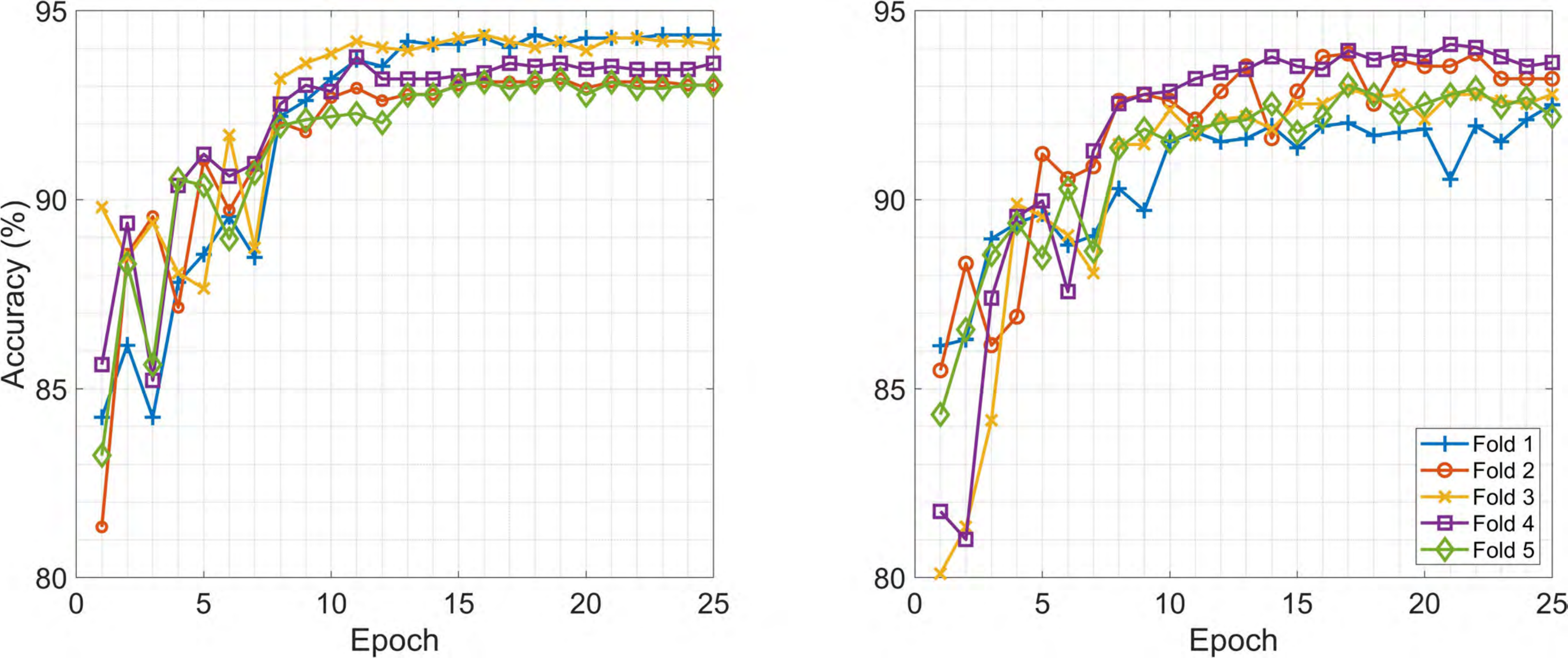}
    \caption{Validation progress curves obtained during 5-fold cross validation for [left] VGG16 and [right] ResNet50.}
    \label{fig:validationAcc}
\end{figure}

In the first training policy, we considered the labelled images by experts and used the $k$-fold cross-validation strategy to evaluate the classifier performance on unseen data. We set the number of folds to five. In this way, the model was trained on four subsets and validated on the remaining subset. The model's performance is tracked by its accuracy and loss obtained by each fold's weighted mean operation, see Figures~\ref{fig:trainingAccLoss} and~\ref{fig:validationAcc}. We also computed each fold's accuracy and reported the average 5-fold results in Table~\ref{tbl:01} with the 95\% confidence interval.

\begin{table}[!htbp]
\centering
\caption{The average classification accuracy and loss of VGG16 and ResNet50 are reported at 95\% confidence intervals.}
\label{tbl:01}
\resizebox{\linewidth}{!}{%
\begin{tabular}{lllcc}
\hline
\multicolumn{3}{l}{\textbf{Method}} & \textbf{VGG16} & \textbf{ResNet50} \\ \hline
\multicolumn{5}{l}{\cellcolor{whitesmoke}5-Fold Cross Validation} \\ \hline
 &  & Accuracy(\%) & $94.61 \pm 0.24$ & $93.64 \pm 0.29$ \\
 & \multirow{-2}{*}{Train} & Loss & 0.130 & 0.180 \\
 &  & Accuracy(\%) & $93.86 \pm 0.54$ & $93.21 \pm 0.62$ \\
 & \multirow{-2}{*}{Test} & Loss & 0.208 & 0.207 \\ \hline
\multicolumn{5}{l}{\cellcolor{whitesmoke}Validating on unconfirmed set} \\ \hline
 &  & Accuracy(\%) & 97.11 & 96.50 \\
 & \multirow{-2}{*}{Train} & Loss & 0.083 & 0.093 \\
 &  & Accuracy(\%) & 90.01 & 87.93 \\
 & \multirow{-2}{*}{Test} & Loss & 0.301 & 0.323 \\ \hline
\end{tabular}%
}
\end{table}

%%%PLACEMENT
\begin{table*}[!htbp]
    \centering
    \caption{Performance of the proposed approach compared to competing methods considering class-wise recall, precision, and F1 score. Here, TP, TN, FP and FN stand for True Positive, True Negative, False Positive and False Negative, respectively.}
    \label{tbl:02}
    \begin{tabular}{lllllllll}
    \hline
    Architecture & \textbf{Number of samples} & \textbf{TP} & \textbf{TN} & \textbf{FP} & \textbf{FN} & \textbf{Precision} & \textbf{Recall} & \textbf{F1 score} \\ \hline
    VGG16 & 6378 & 3318 & 2935 & 46 & 79 & 0.98 & 0.97 & 0.98 \\
    ResNet50 & 6378 & 3270 & 2898 & 94 & 116 & 0.97 & 0.96 & 0.96 \\
    Pataki et al.~\cite{52pataki2021deep} & 7686 & 6041 & 1129 & 362 & 154 & 0.94 & 0.97 & 0.95 \\ \hline
    \end{tabular}%
\end{table*}

Figures~\ref{fig:trainingAccLoss} and~\ref{fig:validationAcc} demonstrate that the VGG 16 architecture achieves a good trade-off between efficiency and number of parameters and outperforms ResNet50 during training and validation in both experimental setups. In theory, any deep CNN architecture can be used with our approach. However, given that deep CNN requires training on large labelled data sets, the use of deeper architectures such as VGG19~\cite{13simonyan2014very}, ResNet101~\cite{12he2016deep} or GoogLeNet~\cite{11szegedy2015going} in our task with a limited number of representative data can reduce generalizability. This statement was validated by comparing the performance metrics of VGG16 and ResNet50 for this study.

Tables~\ref{tbl:02} reports precision, recall and F1 metrics for fine-tuned VGG16 and ResNet50 on the second experiment setup. In this setup, we used labelled mosquito datasets validated by a team of experts (confirmed \textit{Aedes albopictus}) to train and Not-classified tiger mosquito images to test the models. Not-classified cases are tiger mosquitoes submitted by volunteers but not confirmed by expert entomologists of the Mosquito Alert Platform. Although Pataki et al.~\cite{52pataki2021deep} reported a higher True Positive (TP) rate, they predict the positive class at the cost of accepting more non-tiger mosquitoes as tiger mosquitoes, i.e., a significantly higher False Positive (FP) rate.

\begin{figure}[!htbp]
    \centering
    \includegraphics[width=1\linewidth]{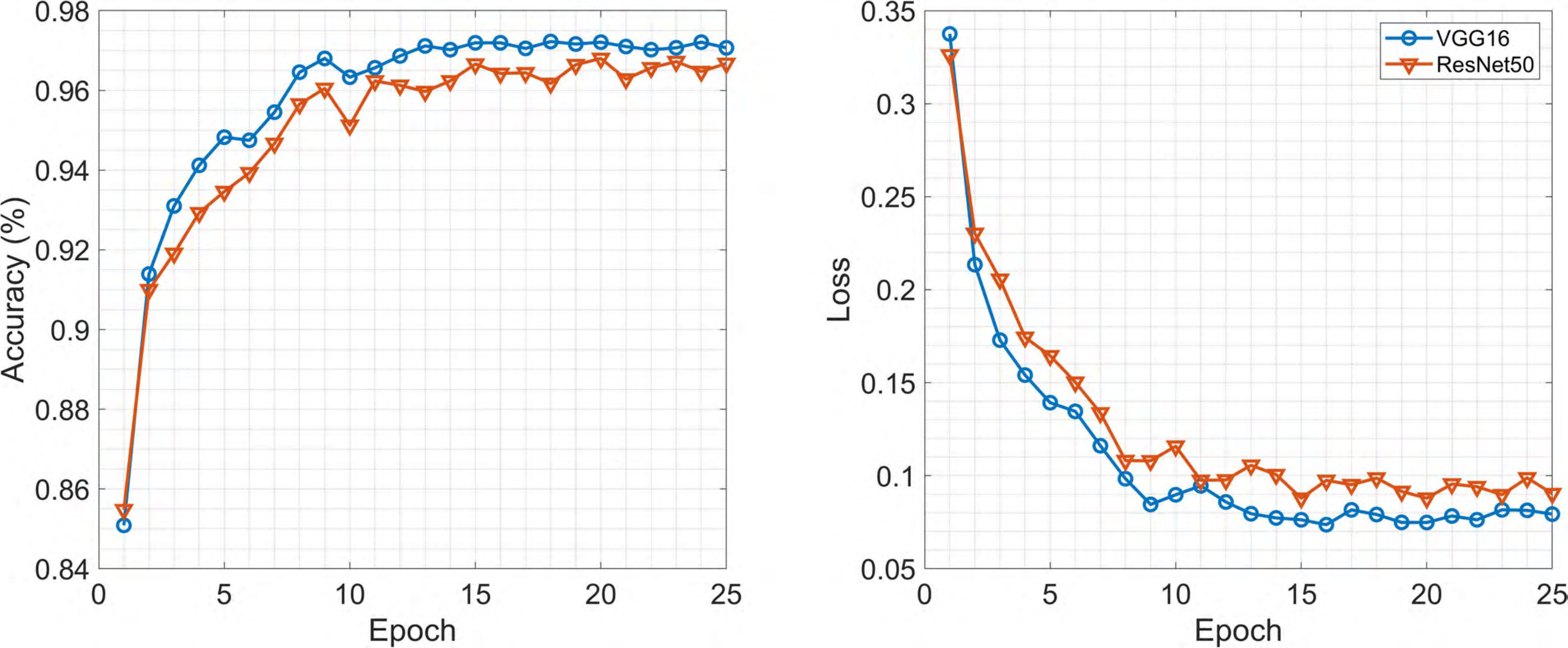}
    \caption{[left] Accuracy and [right] Loss of VGG16 and ResNet50 in training with expert-annotated data and testing with volunteer-annotated images.}
    \label{fig:AccLoss}
\end{figure}

\begin{figure}[!htbp]
    \centering
    \subfigure[]{\includegraphics[width=0.8\linewidth]{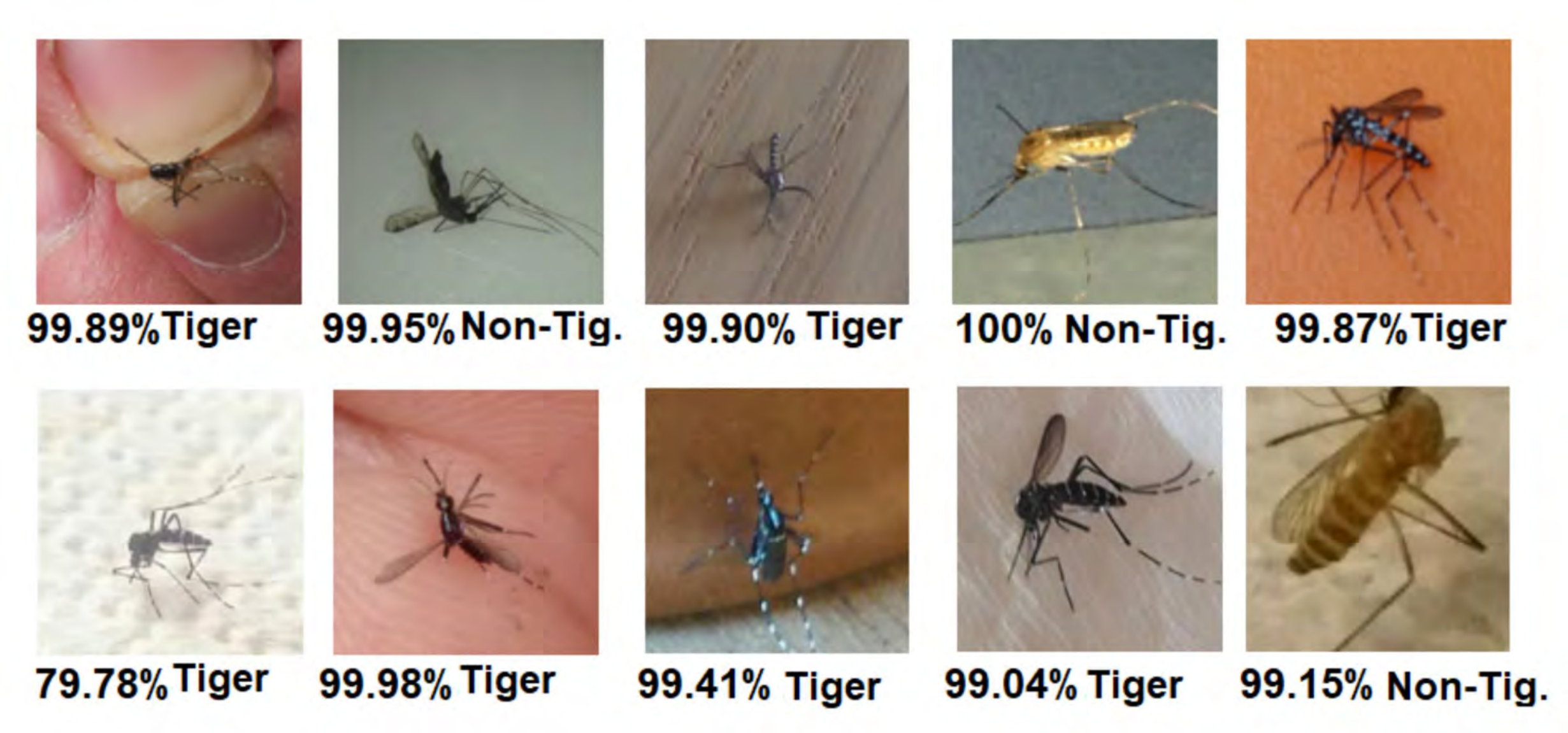}}
    \subfigure[]{\includegraphics[width=0.8\linewidth]{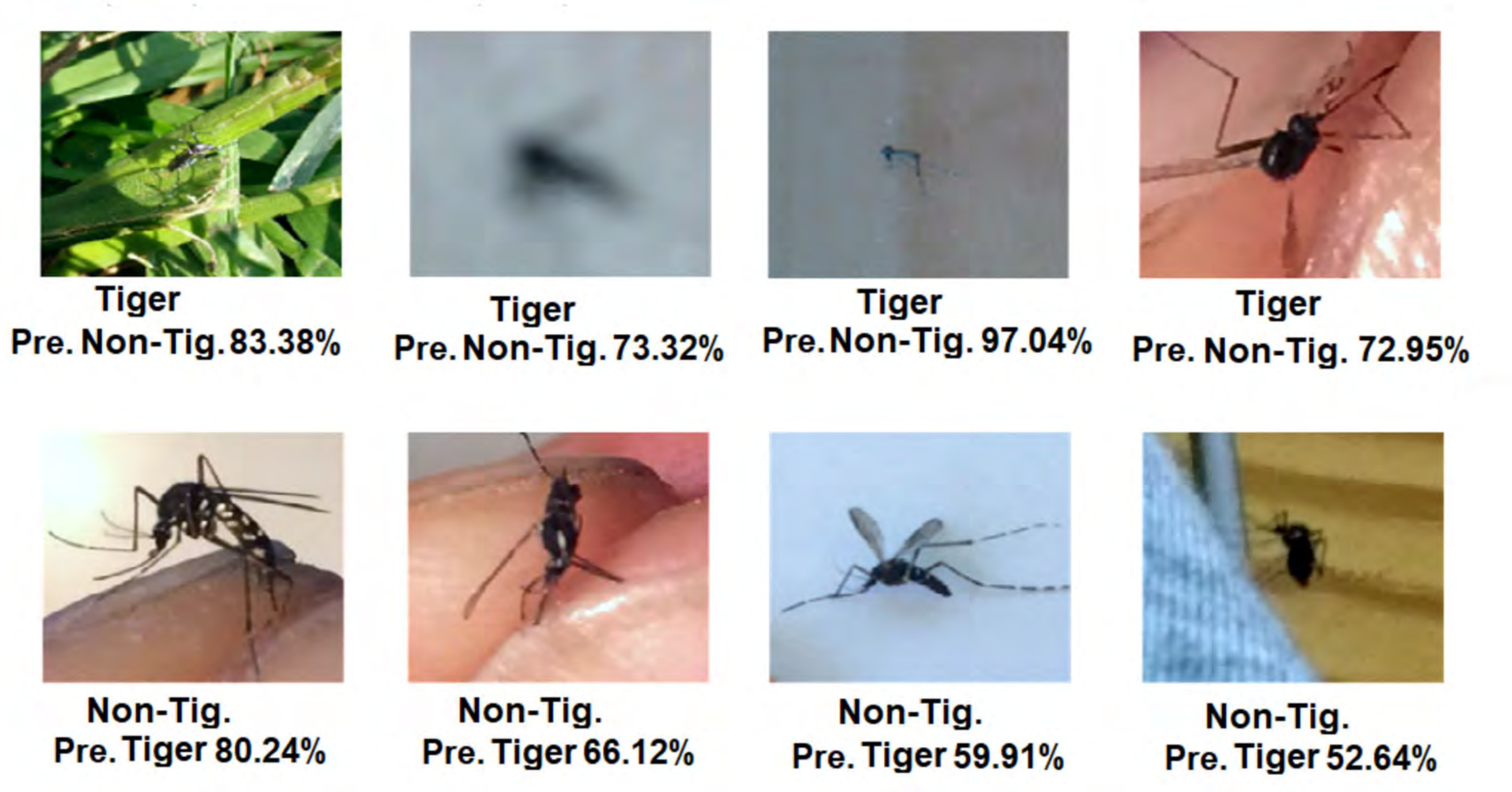}}
    \caption{Sample of images for error analysis, showing (a) correct predictions and (b) prediction with errors.}
    \label{fig:errorSample}
 \end{figure}

Figure~\ref{fig:AccLoss} shows the accuracy and loss for the second set of experiments in which, compared to ResNet50, VGG16 has relatively higher accuracy and less loss. In order to understand the primary source of CNN-induced error, we randomly picked 150 mosquito images (clear and damaged images of the tiger (positive) and non-tiger (negative)) to observe their visualisation prediction scores. The tiger mosquito image is described as a clear image if the image was taken in moderate lighting, the legs are not broken, the white stripes on the legs, abdomen and thorax are visible, and the head is observable. We observed that the model could distinguish between tiger mosquitoes and non-tiger mosquitoes with lower accuracy when the images were of poor quality, distorted, occluded due to pose, or when the mosquito's size was relatively small compared to the image itself, see Figure~\ref{fig:errorSample}. 

\begin{figure}[!htbp]
    \centering
    \subfigure[]{\includegraphics[width=0.8\linewidth]{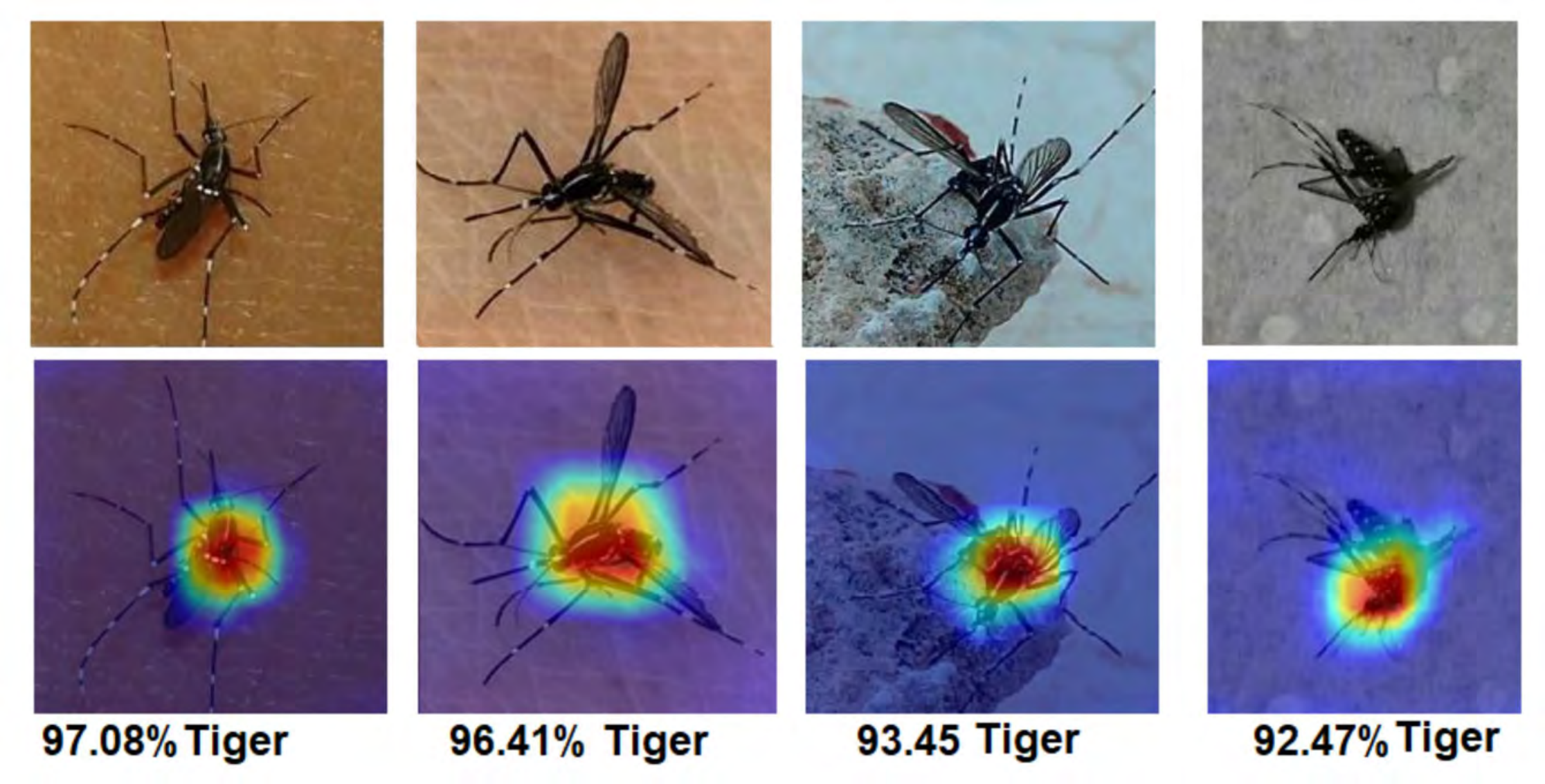} \label{fig:GradCAM1a}}
    \subfigure[]{\includegraphics[width=0.8\linewidth]{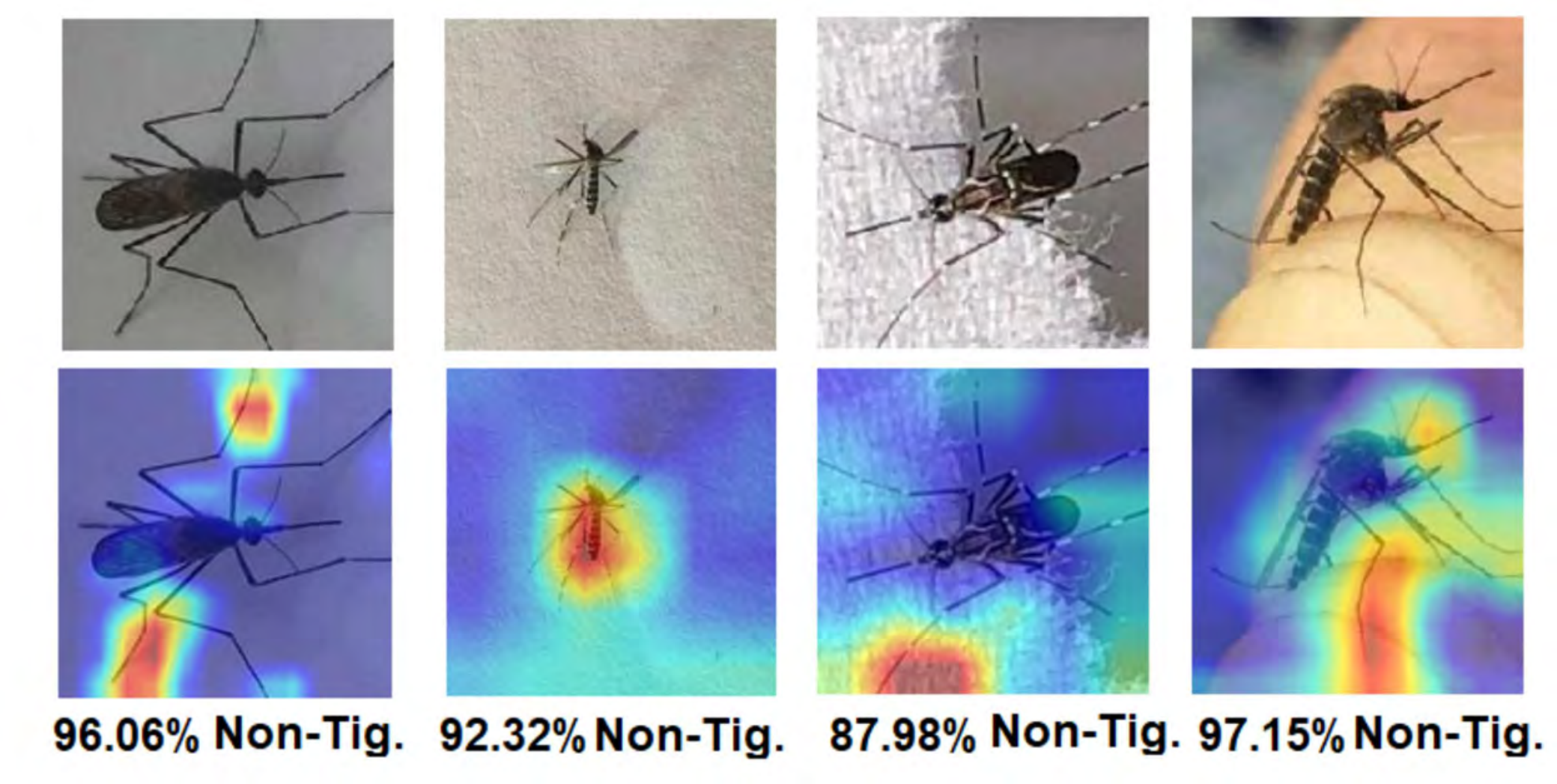}\label{fig:GradCAM1b}}
    \caption{Examples of Grad-CAMs for the (a) tiger and (b) non-tiger mosquitoes species generated from the last convolutional layer.}
\end{figure}

Therefore, training the model with more data not only reduces errors caused by poor image quality but also improves the model's ability to discriminate morphological patterns and, as a result, the mosquito type. The use of a Grad-CAM~\cite{02selvaraju2016grad}, a heatmap that highlights supportive pixels in the image for the classifier, revealed that for an image predicted as tigers (see Figure~\ref{fig:GradCAM1a}), the major key parts strongly activated around the thorax and central part of the mosquito. For non-tiger mosquito input images (see Figure~\ref{fig:GradCAM1b}), however, the legs and abdomen are strongly highlighted and the areas around the wings are slightly highlighted.

\begin{figure}[!htbp]
    \centering
    \subfigure[]{\includegraphics[width=0.8\linewidth]{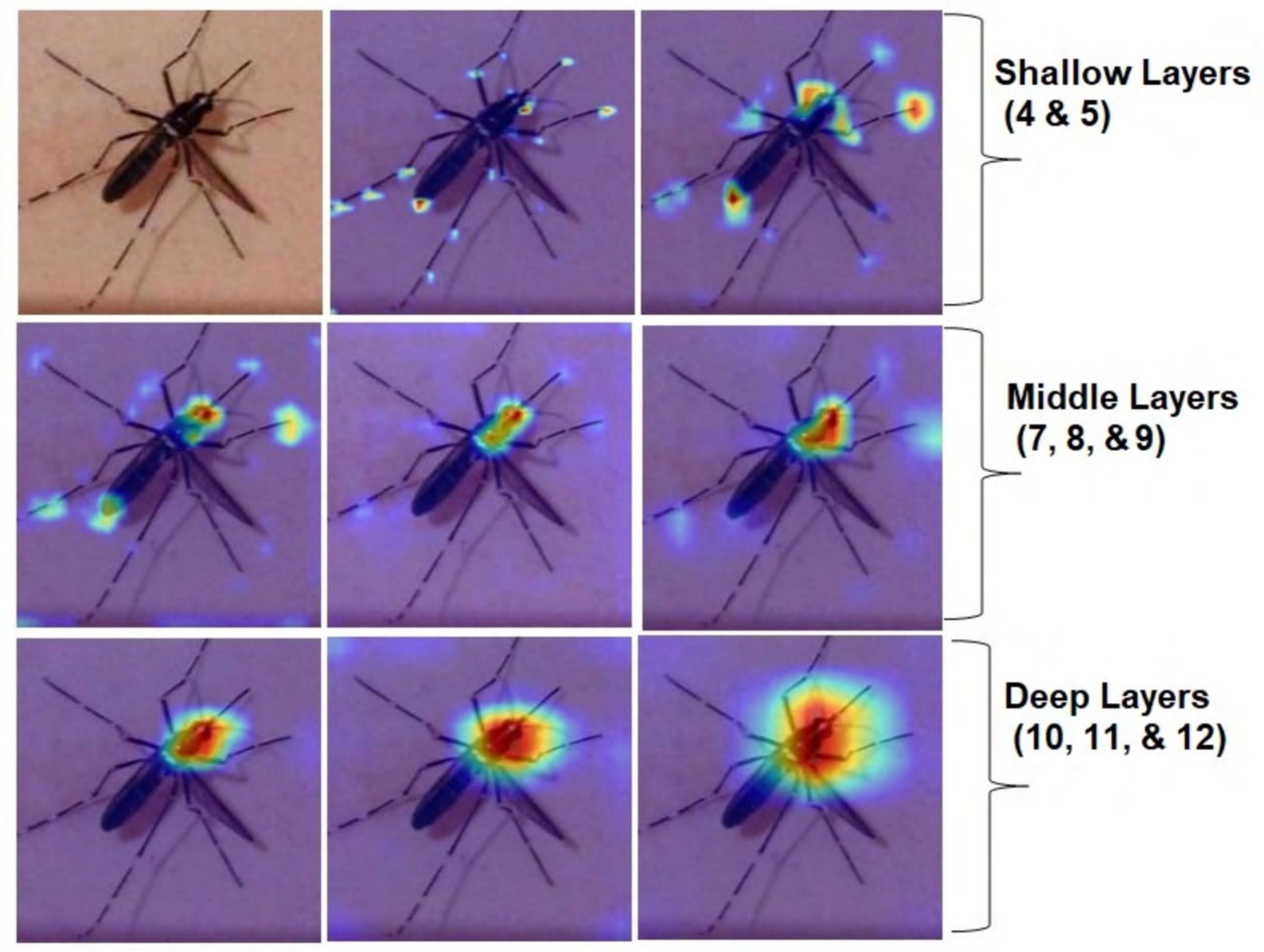}}
    \subfigure[]{\includegraphics[width=0.8\linewidth]{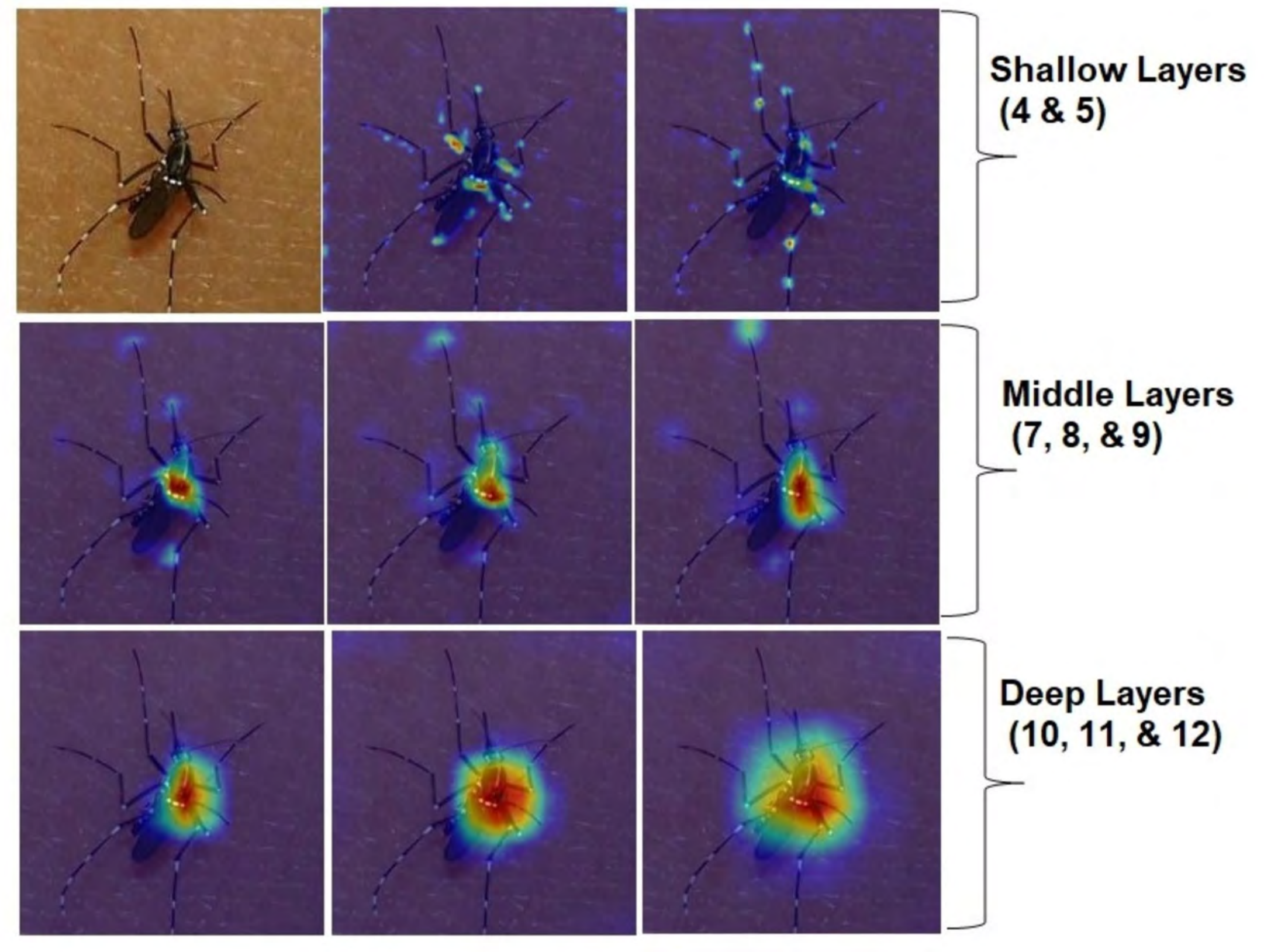}}
    \caption{Visualisation of discriminative regions of images predicted as tiger mosquitoes; the white stripes in the legs and antennae are slightly highlighted in the shallow layers, whereas the thorax is strongly highlighted in the middle and deeper layers.}
    \label{fig:GradCAM2}
\end{figure}

In order to gain insight into the network, we also created the heatmap of the shallow, middle and deep layers of the CNN model. Figure~\ref{fig:GradCAM2} provides an example of Grad-CAM visualisations for a tiger mosquito image at these layers. We observed that the legs, antennae, and proboscis are highlighted as key parts in the shallow layers. The head, thorax, and abdomen are strongly highlighted in the middle layers, while portions of the legs are marginally highlighted. Finally, in the deeper layer, the areas around the the thorax are highlighted that direct the network to determine the class of the input image. The Grad-CAMs generated, specifically, at deeper layers of the trained model indicate that the areas used by the model to detect the type of mosquito are similar to those used by the entomologists.

It should be noted that human experts use a hierarchical structure based on three main features of different weights to achieve a standardised classification. In fact, the observation of white stripes in the thorax leads to immediate identification of \textit {Aedes albopictus} with a high degree of certainty. The stripes in the abdomen and the rear legs are inspected in case of doubt. In mosquitoes, antennae are usually used to classify gender, although there is a very high gender divergence in cephalic areas. In this data set, the majority of tiger mosquito images belong to females. As a result, in addition to the determination of the species, what is discriminated with in shallow layers could be correlated to gender.
 
\begin{figure}[!htbp]
    \centering
    \subfigure[]{\includegraphics[width=0.8\linewidth]{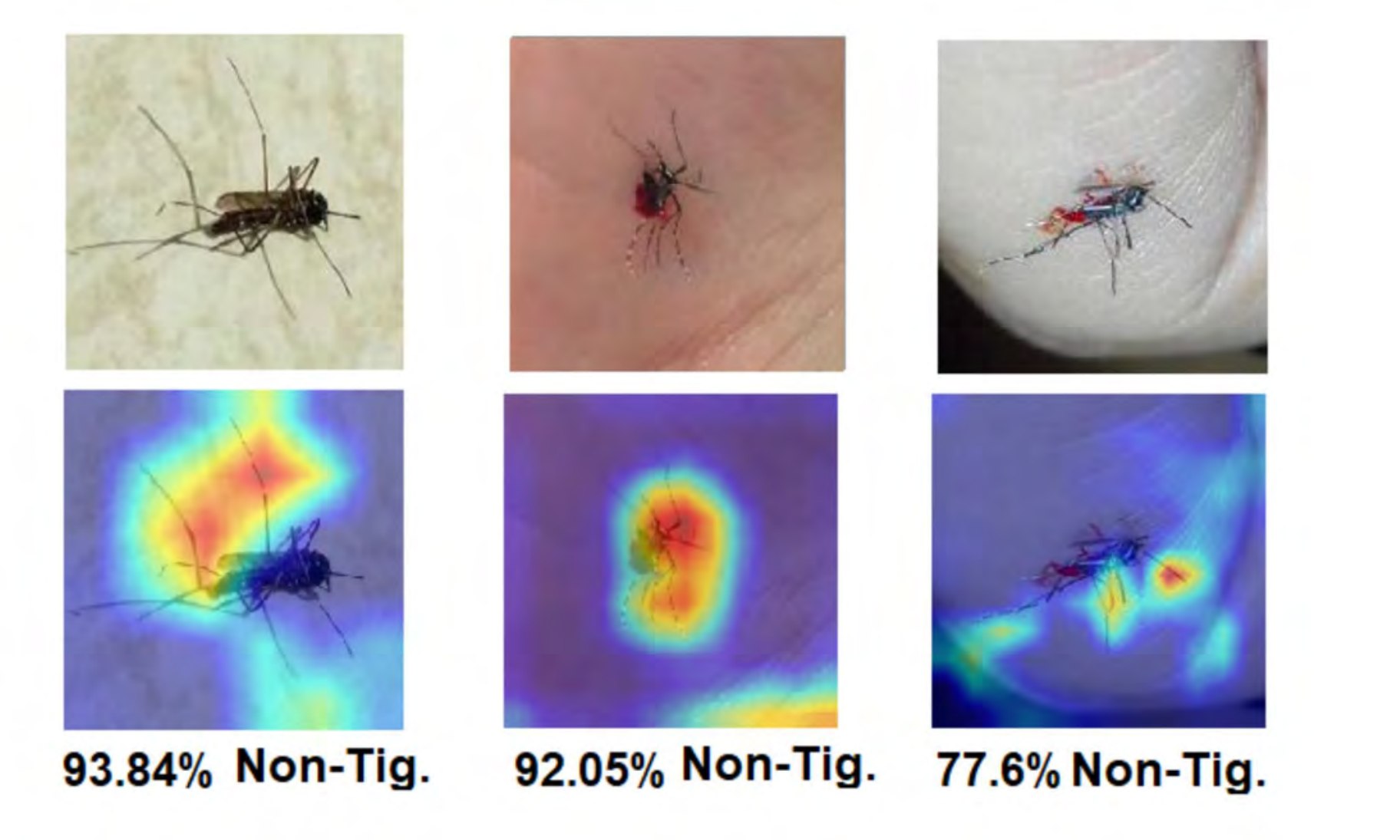}}
    \subfigure[]{\includegraphics[width=0.7\linewidth]{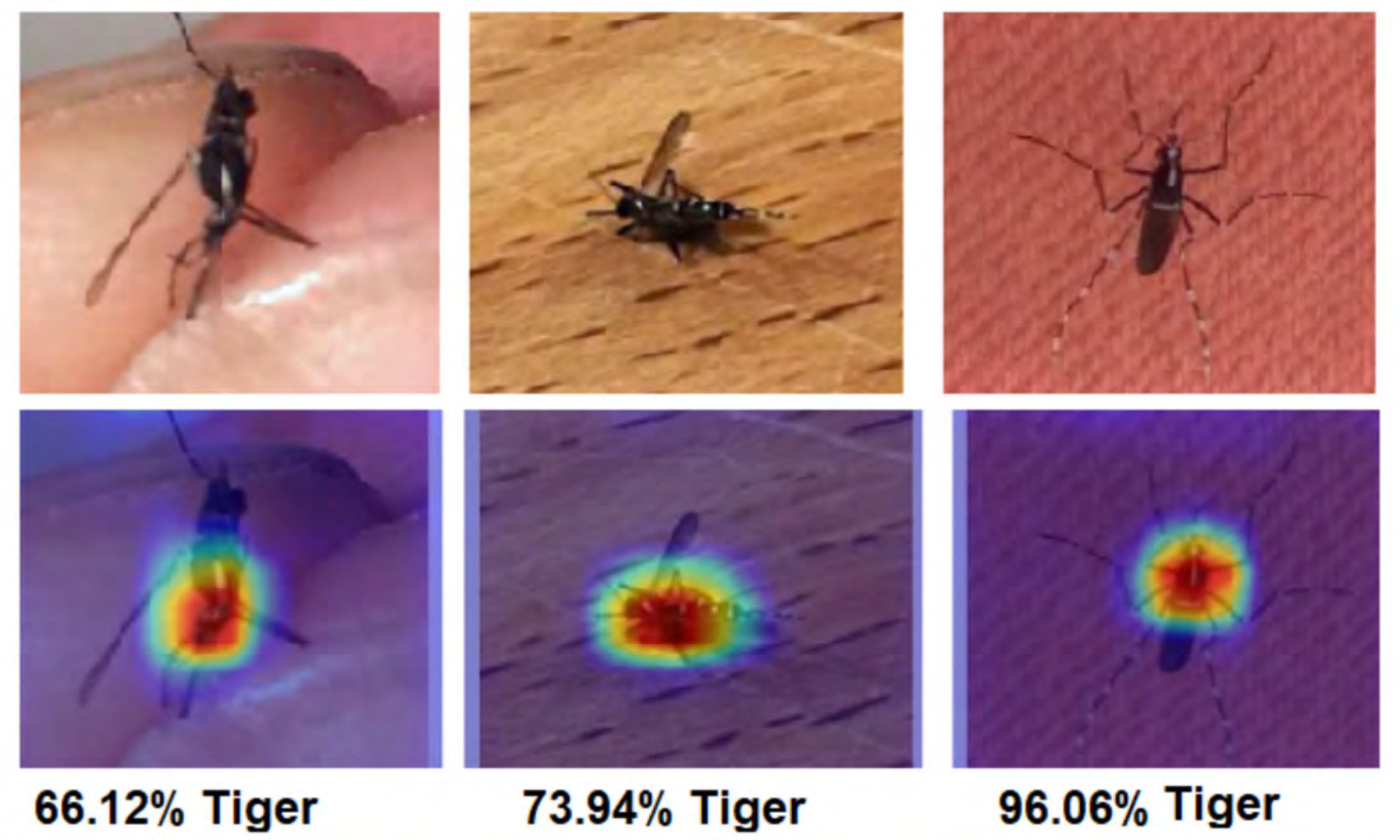}}
    \caption{(a) Examples of tiger input images predicted as non-tiger. (b) Examples of non-tiger input images predicted as tiger. The Grad-CAMs were generated from the last convolution layer.}
    \label{fig:GradCAM3}
\end{figure}

\begin{figure}[!htbp]
    \centering
    \subfigure{\includegraphics[width=1\linewidth]{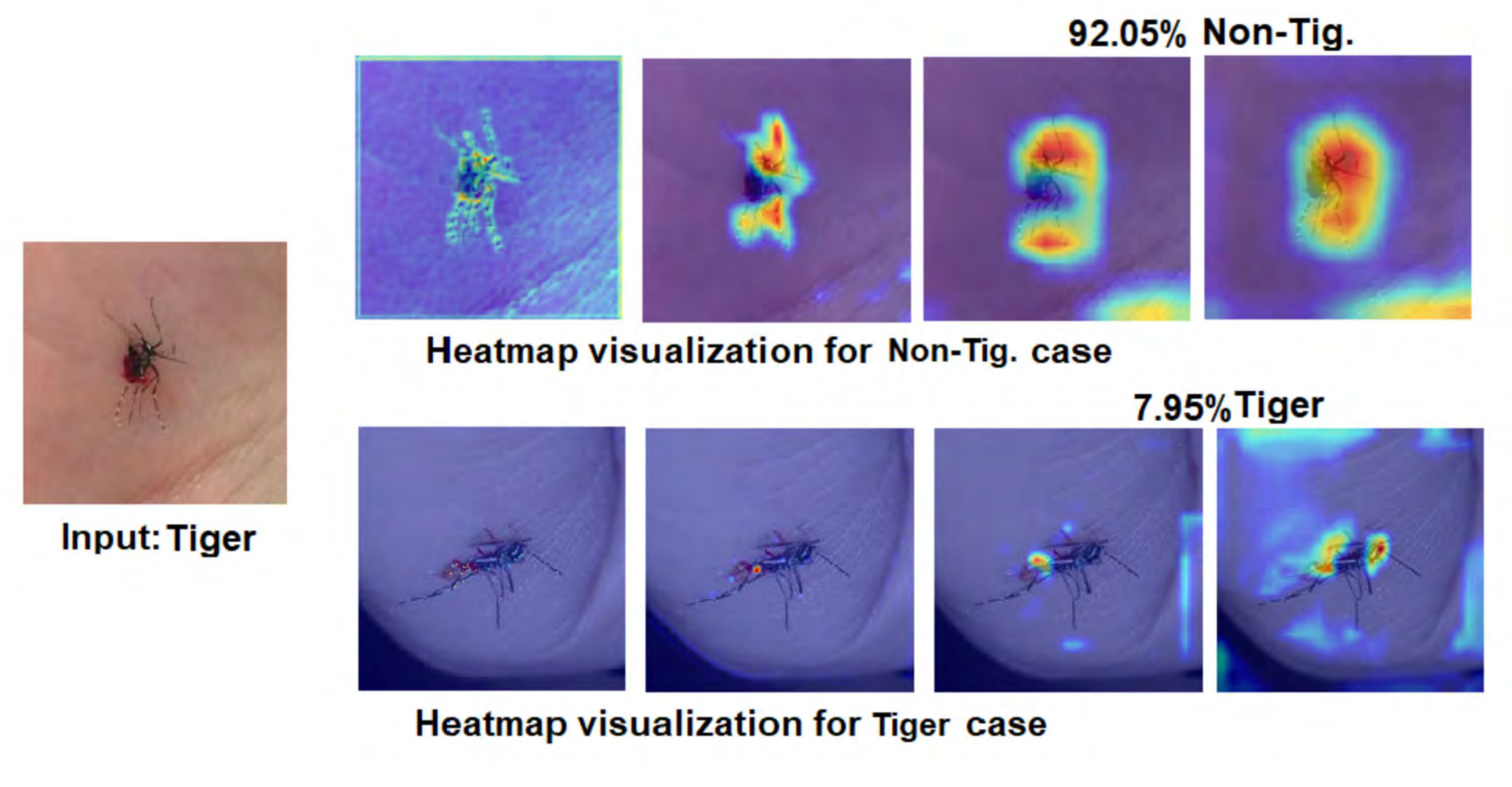}}
    \subfigure{\includegraphics[width=1\linewidth]{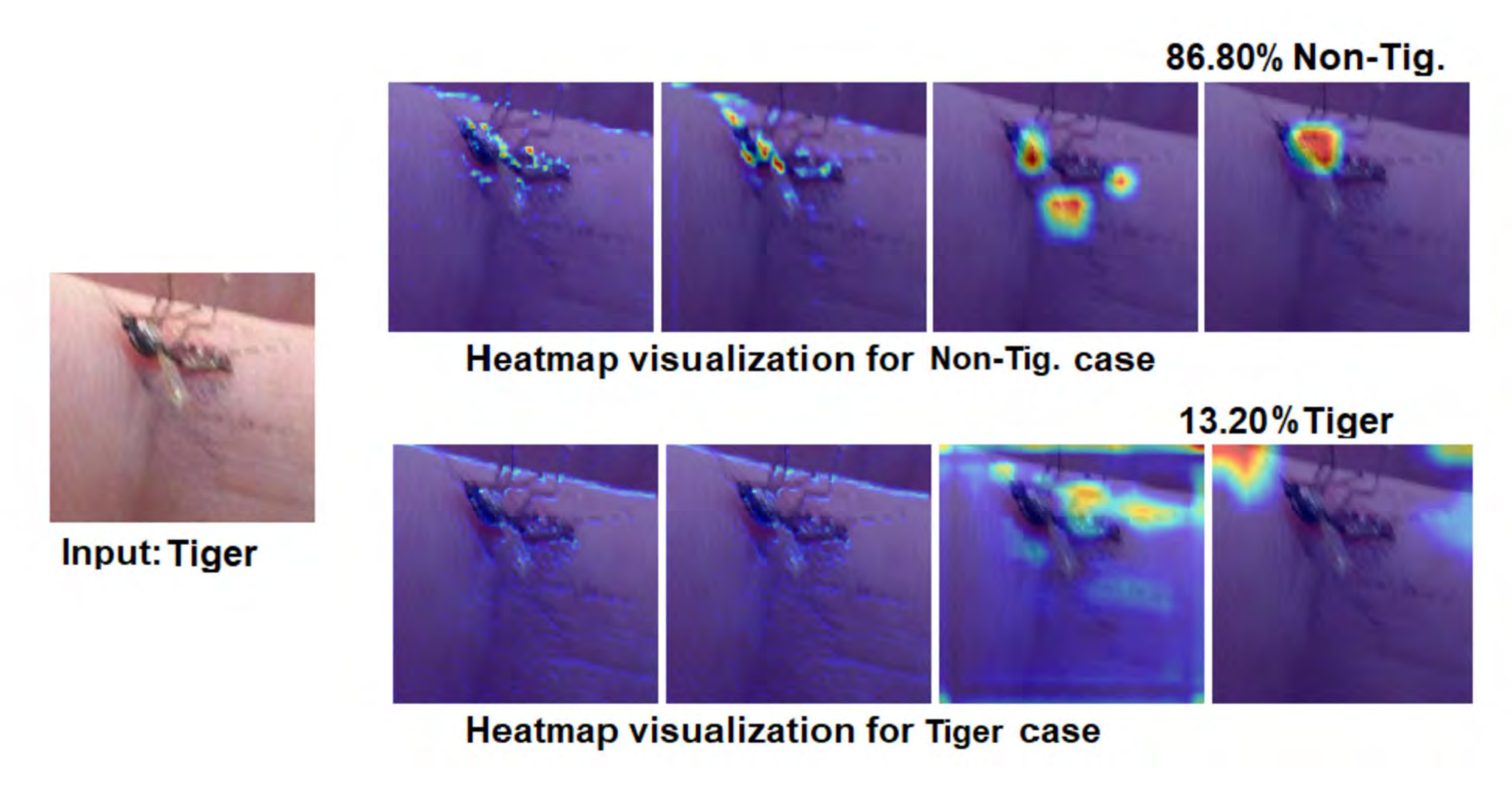}}
    \caption{Visualisation of discriminative regions for tiger images predicted as non-tiger in the shallow, middle and deeper layers and their respective accuracy.}
    \label{fig:GradCAM4}
 \end{figure}

\begin{figure}[!htbp]
    \centering
    \subfigure{\includegraphics[width=1\linewidth]{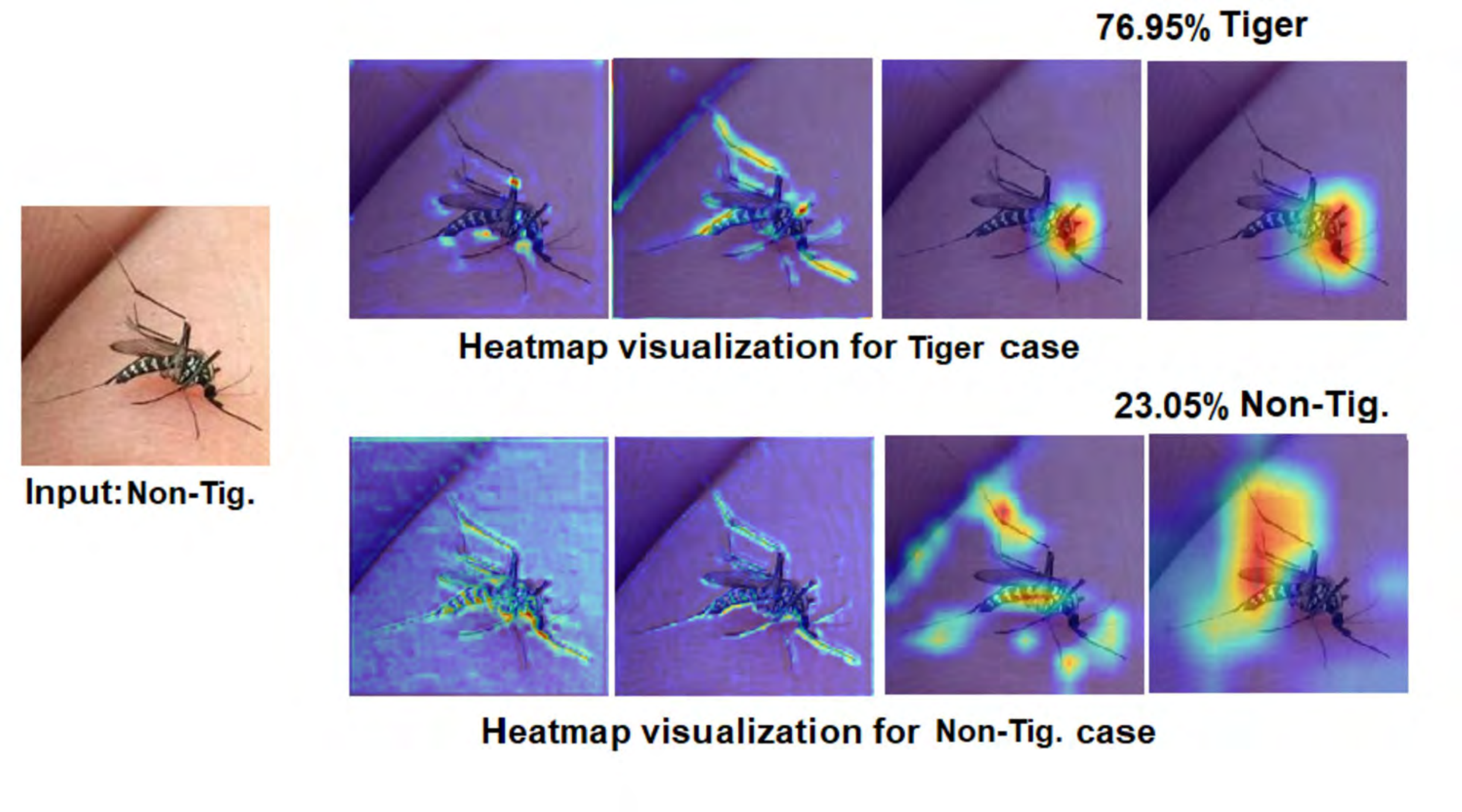}}
    \subfigure{\includegraphics[width=1\linewidth]{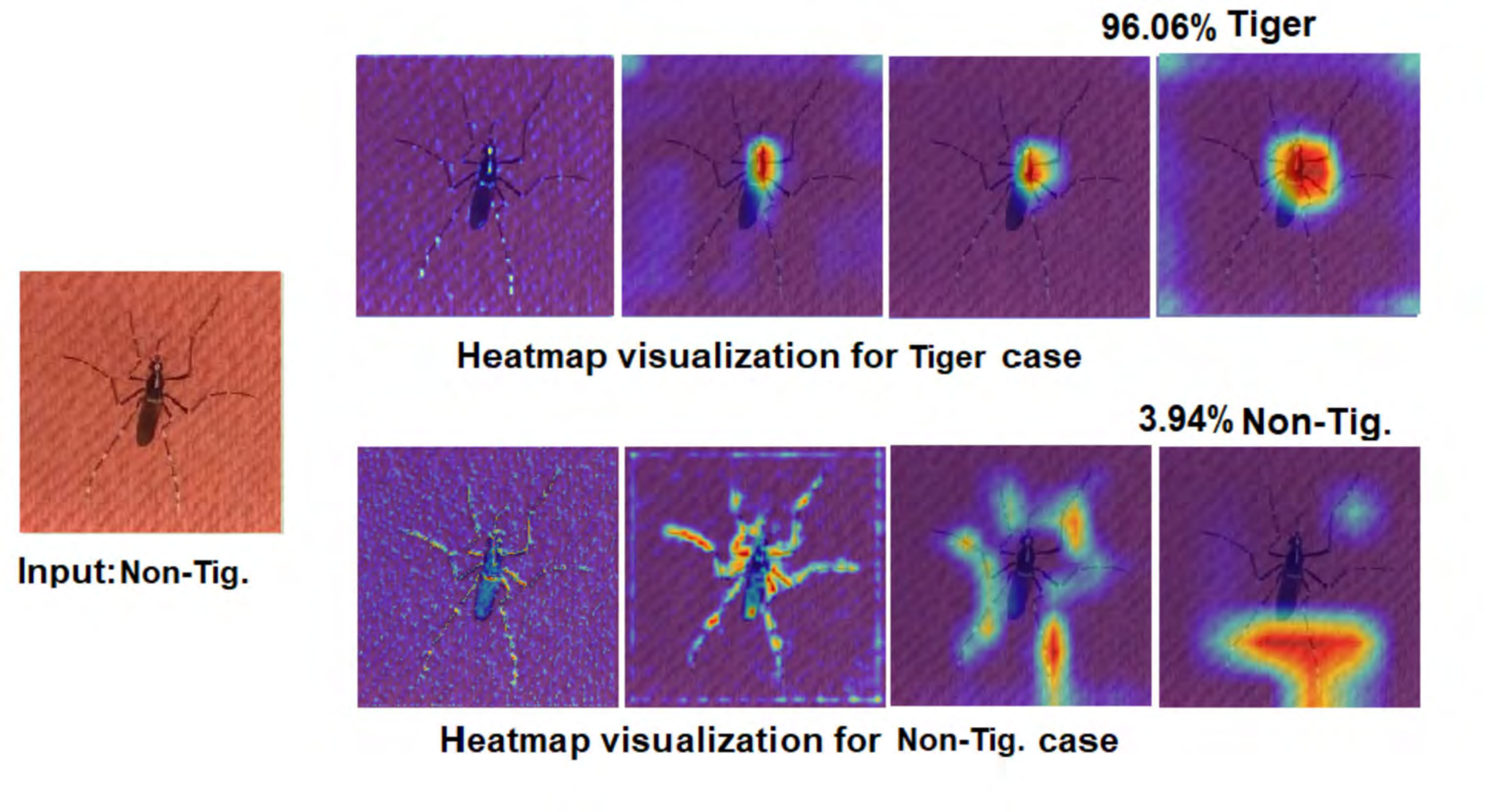}}
    \caption{Visualisation of discriminative regions for non-tiger images predicted as tiger in the shallow, middle and deeper layers and their respective accuracy.}   
    \label{fig:GradCAM5}
\end{figure}

Figure~\ref{fig:GradCAM3} shows examples of misclassified images with Grad-CAMS and their prediction scores. For tiger images that have been misclassified as non-tiger, the morphological patterns of the mosquito body (e.g., legs and thorax) are severely damaged or occluded, see Figure~\ref{fig:GradCAM4}. On the other hand, for non-tiger mosquitoes, that have been misclassified as tiger, the morphological similarity of key parts (e.g., striped legs and abdominal patches) to tiger mosquitoes is the main cause of the error, see Figure~\ref{fig:GradCAM5}.

To sum up, after careful examination of the misclassified tiger images, we found the following two explanations for model confusion:
\begin{enumerate}
    \item The thorax or abdominal parts, which are more informative for the model in the middle and deeper layers for identifying tiger mosquitoes, are damaged or occluded in the input image.
    \item The legs, which are important for the model in shallow layers to locate the mosquito, are either broken or invisible in the input image.
\end{enumerate}

\section{Conclusion} \label{sec:conclusion}
Mosquitoes are the vectors for infectious diseases such as yellow fever, malaria, dengue, Zika, and chikungunya. An estimated 228 million cases of malaria infection with an overall death toll of 405,000 cases have been recorded in 2019~\cite{15worldmalariareport}, requiring efficient mechanisms for early detection and control of mosquito-borne diseases. Mosquito Alert~\cite{01} is an expert-validated citizen science platform launched in 2014 to monitor and control disease-carrying mosquitoes. On this platform, people can upload images and reports of tiger mosquitoes and other invasive mosquitoes to be validated by expert entomologists. As the platform expands, support tools have been provided to assist experts in the validation of the reports submitted.

Pataki et al.~\cite{52pataki2021deep} used mosquito Alert~\cite{01} data set and developed a deep learning model to distinguish between tiger and non-tiger mosquitoes. However, the objective of that work was not to enhance the classification efficiency of the model. Instead, they were mainly concerned with assessing the usability of the labelled images in terms of space and time, as well as the geographic distribution and dynamics of invasive species.  In this study, we developed a vision-based approach to enhance the validation process and characterise morphological features of tiger mosquitoes. We presented a deep convolutional neural network based on VGG16 architecture~\cite{13simonyan2014very} to identify \textit{Aedes albopictus}. This approach achieved a good trade-off between efficiency and number of parameters and outperformed competing approaches in the experimental setups. The data standardisation and fine-tuning of the architecture on Mosquito Alert data set helped to address the small sample size problem and to achieve training and validation accuracy of $94.61 \pm 0.24\%$ and $93.86 \pm 0.54\%$, respectively.

In addition, we studied the discriminative regions of tiger mosquitoes by visualising the target position using the gradient-weighted class activation map. By visualising the discriminative regions, we have identified the white band stripes in the legs, the abdominal patches, the head, and thorax as the main regions used by the classifier to classify the different species. We also observed that most of the classification errors were caused by significant damage to the main areas of the mosquito body.

For future work, as the Mosquito Alert platform is scaling to cover broader geographical areas outside Spain, this study will be expanded to the classification of multiple species of mosquitoes and their breeding sites. We also plan to develop an end-to-end deep learning architecture to learn the most discriminating features from the key regions identified by heatmap visualisation. We plan to incorporate the explainability criteria into the learning loop, combining the heatmaps associated to each class with an anatomical model of the relevant regions of the animal in terms of discriminability. Failure from focusing on the key regions should result on a penalty on the learning loss, promoting classifiers that focus on the a priori knowledge of the classification task.

\section*{Acknowledgement}
This research was supported by ``RTI2018-095232-B-C22'' grant from the Spanish Ministry of Science, Innovation and Universities (FEDER funds), and NVIDIA Hardware grant program. This work was in part financed by EU Horizon 2020 program grant agreement ``VEO'' No. 874735. We also acknowledge the Mosquito team (\url{http://www.mosquitoalert.com/en/about-us/team/}) for their work in keeping the system operative, even in harsh financial times, and most especially the team of volunteer entomology experts that have validated mosquito pictures from Mosquito Alert during the period 2014 to 2019: Mikel Bengoa, Sara Delacour, Ignacio Ruiz, Maria Àngeles Puig, Pedro María Alarcon-Elbal, Rosario Melero-Alcíbal, Simone Mariani, and Santi Escartin. Finally, we would like to thank the Mosquito Alert community (anonymous citizens) who have participated year by year, making all this data collection system worth it.

\bibliographystyle{IEEEtran}
\bibliography{reference}

\begin{IEEEbiography}[{\includegraphics[width=1in,height=1.25in,clip,keepaspectratio]{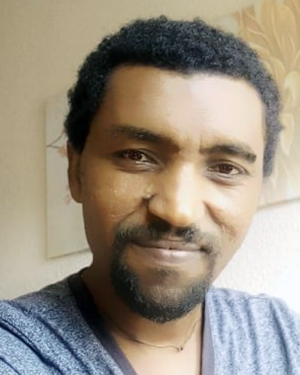}}]{Gereziher Adhane} is currently a PhD student at Universitat Oberta de Catalunya, Spain. He obtained his MSc from Osmania University (India) in 2013/14. His research interests includes deep learning, computer vision and fairness in AI.
\end{IEEEbiography}
\begin{IEEEbiography}[{\includegraphics[width=1in,height=1.25in,clip,keepaspectratio]{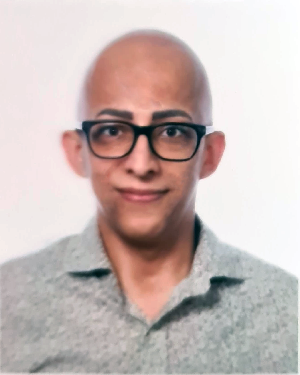}}]{Mohammad Mahdi Dehshibi}
is currently a postdoctoral research fellow at Universitat Oberta de Catalunya, Spain. He obtained the PhD from IAU (Iran) in 2017. He was also a visiting researcher at Unconventional Computing Lab, UWE, Bristol, U.K. He has contributed to over 50 papers published in scientific Journals and International Conferences. His research interests include Affective and Unconventional Computing. 
\end{IEEEbiography}
\begin{IEEEbiography}[{\includegraphics[width=1in,height=1.25in,clip,keepaspectratio]{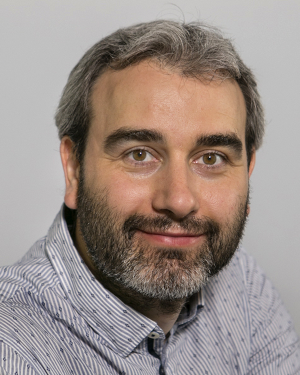}}]{David Masip} is Professor in the Computer Science Multimedia and Telecommunications  Department, Universitat Oberta de Catalunya since February 2007 and Director of the Doctoral School since 2015. He is the director of the Scene Understanding and Artificial Intelligence Lab and member of the BCN Perceptual Computing Lab. He studied Computer Vision at the Universitat Autonoma de Barcelona. He received his PhD in 2005 and was awarded for the best thesis in the Computer Science.
\end{IEEEbiography}
\vfill
% \EOD
\end{document}